\documentclass[10pt,twocolumn,letterpaper]{article}

\usepackage[pagenumbers]{cvpr} % To force page numbers, e.g. for an arXiv version

\usepackage{microtype}

\renewcommand{\paragraph}[1]{\vspace{.5em}\noindent\textbf{#1.}}
\definecolor{cvprblue}{rgb}{0.21,0.49,0.74}
\usepackage[pagebackref,breaklinks,colorlinks,allcolors=cvprblue]{hyperref}
\usepackage[capitalize]{cleveref}
\usepackage{multirow}

\usepackage{xcolor}

\usepackage[table]{xcolor}
\usepackage[normalem]{ulem}
\usepackage[accsupp]{axessibility}  % Improves PDF readability for those with disabilities.
\def\paperID{*****} % *** Enter the Paper ID here
\def\confName{CVPR}
\def\confYear{2026}

\title{PQDT: Pseudo-Query Dual Transformer for Robust Point Cloud Restoration}

\author{
Haoqing Wu$^{1,2}$ \quad
Alexa Nawotki$^{1}$ \quad
Jochen Garcke$^{2,3}$ \\
$^{1}$Mercedes-Benz AG, Germany \quad
$^{2}$University of Bonn, Germany \quad
$^{3}$Fraunhofer SCAI, Germany\\
{\tt\small \{haoqing.wu,alexa.nawotki\}@mercedes-benz.com} \quad
{\tt\small garcke@ins.uni-bonn.de}
}
\begin{document}
\maketitle
\begin{abstract}
Point clouds are a fundamental 3D representation in computer vision, enabling a wide range of perception tasks. However, real-world point clouds often suffer from degradations such as incompleteness, noise, outliers, and irregular density, caused by sensor limitations or occlusions. Recovering clean and detailed shapes from such degraded data is crucial for downstream applications. While existing learning-based methods achieve progress on individual tasks like completion or denoising, they typically rely on global bottleneck features, which lose fine-grained geometry and remain sensitive to varying input quality. We propose a unified 3D restoration network that directly takes point clouds as input and adaptively reconstructs high-quality geometry under diverse degradation scenarios. At the core of our approach is a Pseudo-Query module, implemented within a Transformer backbone, which reformulates geometric translation into two cooperative stages to enhance structural clarity, robustness, and local detail preservation. Extensive experiments on curated benchmarks demonstrate that our approach surpasses state-of-the-art performance in general 3D restoration. It effectively handles complex combinations of completion, deformation, and denoising degradations. With this work, we provide a novel unified, point-only backbone for robust 3D restoration, enabling more versatile 3D perception.
% \HW{maybe:We present a unified, point-only backbone for general 3D restoration, to our knowledge the first to demonstrate strong performance across completion, deformation, and denoising tasks.}
\end{abstract}    
\section{Introduction}
\label{sec:intro}
Point clouds as a very compact and efficient 3D representation have been widely used in many computer vision applications.
With the increasing data collected by 3D sensors, 
we are witnessing a surge of interest in 3D perception tasks 
\cite{qi2017pointnet,qi2017pointnet++,zhao2021point,wu2022ptv2,wu2024ptv3}.
However, we are also facing many challenges caused by the inevitable data quality issues,
such as sparse, incomplete shapes with noise and outliers caused by sensor limitation and self- or inter-occlusion.
Therefore, recovering clean and detailed 3D shapes from degraded point clouds is of great importance for various downstream tasks.
Since input point clouds are typically corrupted by multiple types of degradation simultaneously,
such as missing or distorted points, local or global noise, and varying point densities,
we focus on general 3D restoration where multiple fundamental 3D tasks,
including completion, deformation, and denoising, are combined.
Such 3D restoration capability is crucial for reliable perception in real-world scenarios, 
including autonomous driving, robotic manipulation, augmented and virtual reality (AR/VR) reconstruction, or reverse engineering, where 3D observations are commonly degraded by sparsity, noise, or occlusion artifacts.
\begin{figure}[t]
    \centering
    \includegraphics[width=0.99\linewidth]{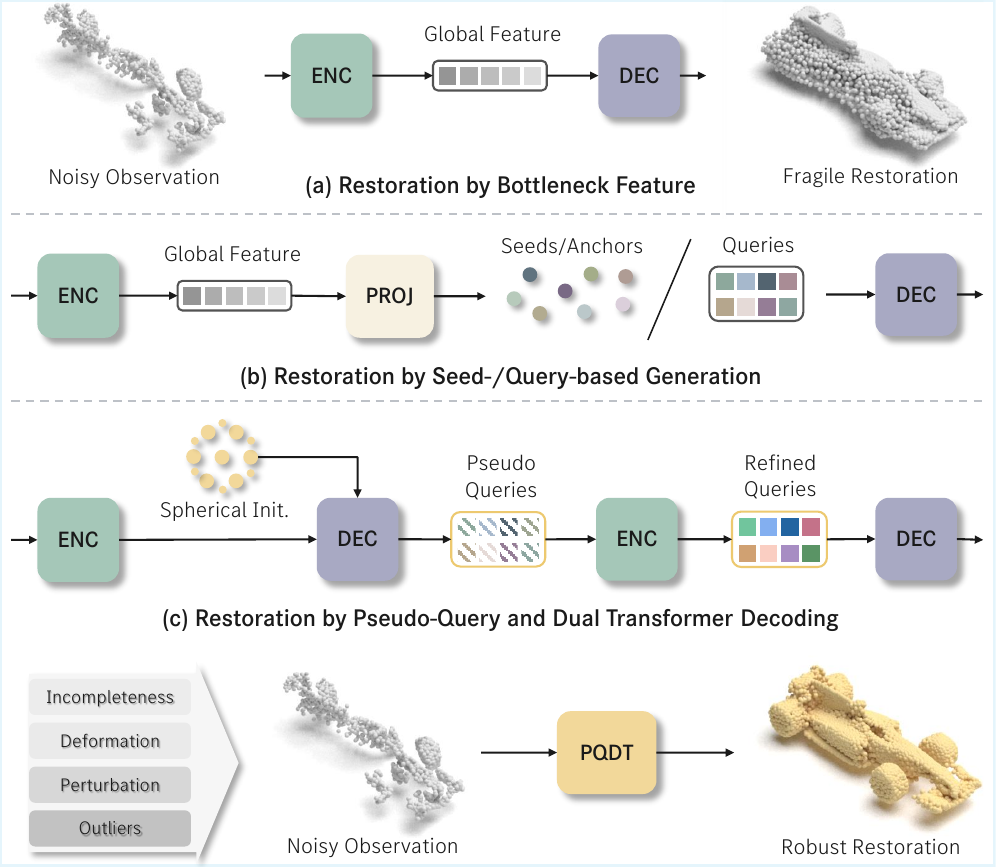}
     \caption{Comparison of point cloud restoration paradigms. 
     (a) Bottleneck feature–based methods directly decode the shape from a global latent code. 
     (b) Seed-/query-based methods generate anchors or queries from the global feature for coarse-to-fine decoding. 
     (c) Our PQDT introduces geometry-aware pseudo-queries as auxiliary entities and dual transformer decoding, 
     achieving more faithful restorations with superior geometric fidelity.}
     \label{fig:teaser}
     \vspace{-10pt}
  \end{figure}
  
Recent advances in deep learning have introduced numerous methods for point cloud completion. 
Most of these adopt an encoder-decoder framework that extracts a global feature from partial inputs 
and then reconstructs the complete shape~\cite{yuan2018pcn,xiang2021snowflakenet,huang2020pf,wen2020point}, 
thereby addressing the permutation invariance of point sets. 
However, this paradigm often sacrifices fine-grained geometric details due to global feature pooling, as shown in \cref{fig:teaser}(a).
To alleviate this issue, several approaches \cite{yu2021pointr,chen2023anchorformer} utilize coarse-level point-wise features to produce intermediate proxies or anchors, 
which help preserve local geometric details and lead to improved reconstruction quality, as shown in \cref{fig:teaser}(b).
Despite these advances, such methods remain sensitive to partial point clouds that may contain noise, outliers, 
or locally or globally distorted geometry, and thus have varying levels of reliability. 
Since the employed reconstructions are decoded directly from bottleneck features or seed points derived from the input, 
their performance can degrade significantly under imperfect observations. 
In real-world scenarios, where captured point clouds rarely align cleanly with the ground-truth geometry \cite{kong2023robo3d}, 
a more robust and adaptive model is needed to handle diverse input qualities.

To this end, we propose a novel transformer-based backbone with auxiliary intermediate entities, termed pseudo-queries, 
designed to recover high-quality shapes under diverse input degradations. 
We decompose the transformer into two complementary and functionally distinct stages: 
\begin{enumerate}[(i), labelwidth=!]
   \item an observation-guided stage, which generates pseudo-queries by broadly attending to encoded inputs, stabilizing coarse geometry from distorted observations;
    \item a prior-guided stage, which refines these queries under stronger influence of learned shape priors, selectively balancing prior knowledge with input evidence to enforce geometric consistency and structural plausibility.
\end{enumerate}
In both stages, our geometric-embedding (GE) transformer encoders capture local and global contextual information, forming the basis for adaptive query generation 
that dynamically selects representative tokens from the input points and the learned pseudo-queries. 
The transformer decoder with our GE enhancement then transforms these learned representations into decoded point proxies, which summarize the underlying surface and are subsequently refined through a coarse-to-fine expansion, progressively restoring local structures and yielding consistent, detailed reconstructions.

In summary, our main contributions are threefold:
\begin{itemize}
  \item We propose a novel pseudo-query dual transformer (PQDT) framework, 
  which effectively unifies completion, denoising, and deformation within a single model. 
  \item 
  We craft a powerful 3D backbone built upon GE self-attention and dynamic query selection, 
  which jointly enable adaptive feature learning that balances input consistency and learned geometric priors.
  \item We construct and present three new datasets covering diverse degradation types, 
  and facilitate standardized evaluation of general 3D point cloud restoration.
\end{itemize}
Extensive experiments on both existing datasets and our new benchmarks demonstrate that 
  our method achieves state-of-the-art performance across multiple tasks and degradation scenarios, 
  significantly surpassing prior approaches in robustness and reconstruction quality.
By enabling accurate recovery of detailed geometry from degraded inputs, it enhances the robustness and applicability of 3D vision systems across diverse domains. Code and data are available at \url{https://github.com/ins-uni-bonn/PQDT}.

\begin{figure*}[t]
    \centering
    \includegraphics[width=0.99\linewidth]{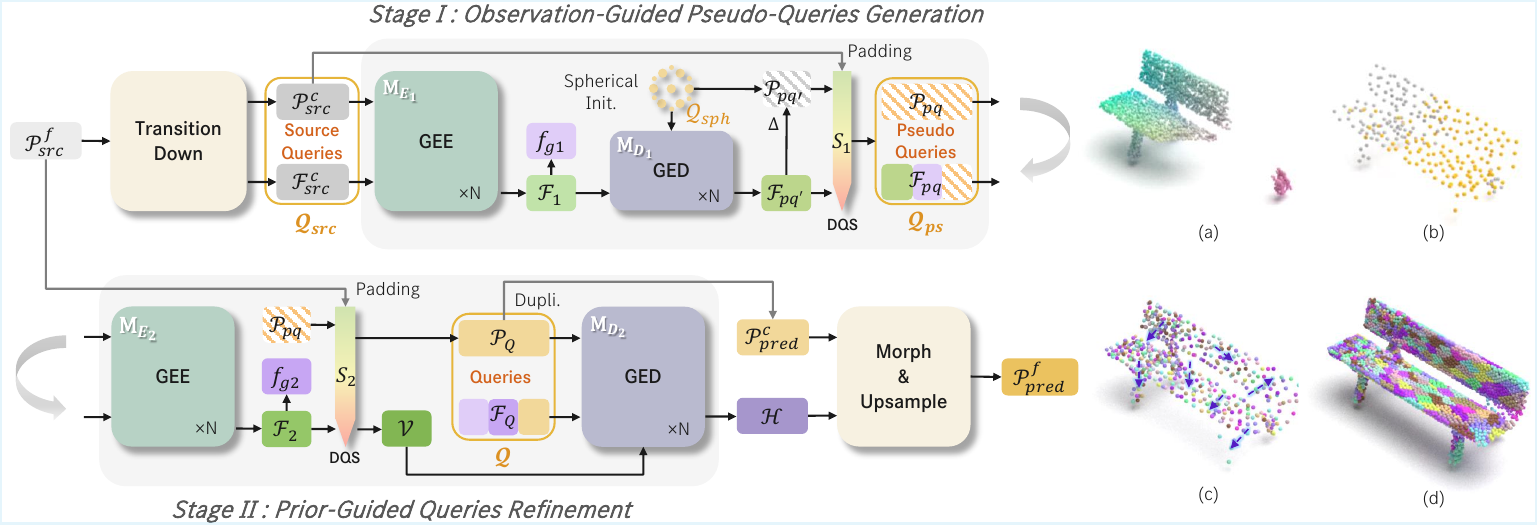}
    \caption{\textbf{Overview of PQDT}. 
    Given an incomplete and noisy input point cloud, 
    local features are extracted via a transition-down module and translated by a dual transformer. 
    Stage I generates pseudo-queries $\mathcal{Q}_{ps}$ through an encoding–decoding process, providing observation-guided query initialization.
    Stage II further refines $\mathcal{Q}_{ps}$, producing consolidated queries 
    $\mathcal{Q}$ to predicted proxies $\mathcal{H}$ that are upsampled to generate fine-level points.
    Blocks with identical colors indicate the same feature/entity reused across modules.
    (a) Input partial point cloud $\mathcal{P}_{src}^f$; (b) Coordinates of pseudo-queries $\mathcal{P}_{pq}$, 
    gray points are downsampled from input, orange points are newly predicted seeds; 
    (c) Coarse-level prediction obtained by morphing (blue arrows) query coordinates $\mathcal{P}_{Q}$ guided by $\mathcal{H}$;
    (d) Fine-level prediction with high geometric consistency.
    }
    \label{fig:model}
    \vspace{-10pt}
\end{figure*}
\section{Related Work}
\label{sec:relatedwork}

Much of the research into 3D shape restoration focuses on completion tasks. 
These tasks commonly arise in real-world scenarios. 
Some early attempts based on 
handcrafted features \cite{berger2014state,hu2019local,mitra2013symmetry,sarkar2017learning} or 
template matching \cite{li2015database,pauly2005example,kim2013learning,sung2015data} 
require a large amount of prior knowledge and are not robust enough to handle new and complex shapes.
Deep learning based methods have shown great potential in 3D geometric processing.
The traditional voxel-based methods \cite{xie2020grnet,dai2017shape,yang20173d,han2017high,stutz2018learning} 
adopting 3D convolutional networks have the limitation of 
high computational cost and low resolution. Recent works focus on point cloud-based shape completion, 
which is more efficient and scalable due to the sparsity of point clouds as 3D representations.

\paragraph{Point Cloud Based 3D Shape Completion} 
The pioneering point-based feature learning networks 
PointNet and PointNet++ \cite{qi2017pointnet,qi2017pointnet++}, 
laid the foundation for deep learning on point clouds. 
The first learning-based framework for point cloud completion is the point completion network (PCN)~\cite{yuan2018pcn}, which reconstructs complete shapes from partial inputs through an encoder-decoder architecture that leverages global features.
To better recover the fine details of the geometry, works like \cite{huang2020pf,tchapmi2019topnet,wen2020point,wang2020cascaded,huang2021rfnet} 
formulate a hierarchical decoder structure and produce the multi-scale generation progressively.
Methods like SnowflakeNet \cite{xiang2021snowflakenet} and SeedFormer \cite{zhou2022seedformer} 
utilize the vector self-attention \cite{zhao2021point} to capture the contextual information 
and preserve the local details in the upsampling process.
Anchorformer \cite{chen2023anchorformer} leverages pattern-aware discriminative anchors 
to replace the global feature projection as sparse points prediction.
By adopting the strength of transformers in sequence-to-sequence modeling, 
PoinTr \cite{yu2021pointr} first converts the completion task to a set-to-set translation problem,
followed by various transformer-based successors \cite{Yu2023AdaPoinTrDP,li2023proxyformer,duan2024t}.
Moreover, beyond single partial point cloud as limited input information, 
approaches like \cite{hu2019render4completion,zhang2021view,gong2021me,wu2023leveraging,zhu2023svdformer,du2025superpc,xu2024explicitly} 
explore the single- or multi-view images to fuse the 2D-3D features for better shape understanding.
Also as multi-modal exploration, works such as \cite{zhou2025position,kasten2023point} 
propose text-guided 3D networks to achieve shape completion with high-level semantic constraints and compensate the missing parts with reasonable structures.

\paragraph{Multi-task Point Cloud Restoration}
In real-world scenarios, the acquired point clouds are usually incomplete and noisy simultaneously.
The common strategy is to handle the single tasks separately with dedicated networks.
Beside the shape completion networks mentioned above, point cloud denoising methods with 
non-learning-based \cite{zhang2019point,zheng2018rolling,fleishman2005robust,hu2020feature} 
and learning-based \cite{rakotosaona2020pointcleannet,pistilli2020learning,mao2022pd,de2023iterativepfn,mao2024denoising} 
manners have been widely studied.
3D shape deformation is another important task in 3D geometric processing,
which aims to modify the source geometry to fit the target shape.
Works like \cite{wang20193dn,uy2021joint,groueix20183d,uy2020deformation} propose mesh deformation networks 
with 2D/3D guidance, which directly predict the vertex offsets so that the topology of the source mesh is preserved.
Given an incomplete shape, without remeshing the source shape, the details in the missing regions cannot be well recovered.
AdaPoinTr \cite{Yu2023AdaPoinTrDP} proposes a noise resistant point cloud completion network to 
encounter the difficulty of incompletion and noise simultaneously. An auxiliary denoising task is designed to 
enhance robustness in the decoder stage.
SuperPC \cite{du2025superpc} presents a diffusion-based \cite{ho2020denoising,luo2021diffusion} generative framework 
with 2D/3D inputs as conditions of the denoising process, 
which can handle the completion, upsampling, denoising and colorization tasks in a unified manner.

Our approach provides a unified framework focusing on the 3D point cloud restoration tasks, 
which covers most common degradation scenarios in real-world applications.

\section{Method}
\label{sec:method}
To effectively capture the geometric structure of defective input point clouds,
we propose a transformer-based architecture that employs auxiliary pseudo-queries as a noise-resistant backbone.
Overall, the proposed framework adaptively balances input fidelity and learned priors, 
enabling the network to handle incomplete, noisy, or unevenly distributed point clouds.
Next, we give an overview of our Pseudo-Query Dual Transformer (PQDT) framework 
in \cref{fig:model}, followed by the description of each submodule. 

Given an incomplete and noisy input point cloud $\mathcal{P}_{src}^f$, 
we first extract the local feature $\mathcal{F}_{src}^c$ with a light-weighted transition-down module, 
and let the coarse level point-wise feature be translated via several novel geometric-embedding encoder (GEE) blocks. 
With spherical points serving as the static query initialization, we employ our geometric-embedding decoder (GED) blocks to build up a DETR-like decoder \cite{carion2020end}. 
Intermediate entities,
serving as an observation guidance, are generated after the first decoding stage and further filtered by a dynamic query selection (DQS) module to obtain $\mathcal{P}_{pq}$ and $\mathcal{F}_{pq}$, which we call pseudo-queries. 
Instead of being used directly for decoding (hence the term pseudo-queries), 
these entities will be further refined through a second encoding stage.
After another DQS step, the consolidated queries are passed to the decoder.
The query coordinates are morphed and upsampled to the fine level prediction base on the decoded point-wise features.

\subsection{Transformer with Auxiliary Pseudo-Query}
As the overall structure in \cref{fig:model} shows, with a two-step encoding scheme we aim to enhance robustness in sequence translation: 
the auxiliary pseudo-queries act as noise-resilient anchors, 
while the DQS mechanism ensures that only the most representative features 
are retained throughout the query generation.

Inspired by point transformer designs \cite{yu2021pointr,Yu2023AdaPoinTrDP}, 
we adopt a sequence-to-sequence generation paradigm based on coarse-level point features, 
referred to as point proxies in \cite{yu2021pointr}.
With the given set of coarse level point proxies as source queries $\mathcal{Q}_{src}=\{\mathcal{P}_{src}^c,\mathcal{F}_{src}^c\}$,
the generation of queries can be formulated as a two-step process:
\begin{align}
    {\mathcal{Q}_{ps}} &= \text{S}_1(\text{M}_{D_1}({\mathcal{Q}_{sph}}, {\text{M}_{E_1}}(\mathcal{Q}_{src})), \mathcal{P}_{src}^c),
    \label{eq:enc1}\\
    \mathcal{V},\mathcal{Q} &= \text{S}_2({\text{M}_{E_2}}(\mathcal{Q}_{ps}), {\mathcal{P}_{src}^f}).
    \label{eq:enc2}
\end{align}
% \begin{equation}
%     {\mathcal{V}_{ps}},{\mathcal{Q}_{ps}} = \text{S}_1({\text{M}_{{E_1}}}(\mathcal{F})),
%     \label{eq:enc1}
% \end{equation}
% \begin{equation}
%     \mathcal{V},\mathcal{Q} = \text{S}_2({\text{M}_{{E_2}}}(\mathcal{V}_{ps}),{\mathcal{Q}_{ps}}, {\mathcal{Q}_{in}}).
%     \label{eq:enc2}
% \end{equation}
Here, $\text{M}_{E_1}$, $\text{M}_{E_2}$, and $\text{M}_{D_1}$ denote GEE and GED, 
and $\text{S}_1$ and $\text{S}_2$ are DQS modules that filter non-representative point proxies and queries.
The auxiliary pseudo-queries $\mathcal{Q}_{ps}$ are first decoded from the outputs of $\text{M}_{E_1}$ and 
the static spherical point queries $\mathcal{Q}_{sph}$ in a DETR-like manner. 
The resulting seed proxies are then combined with $\mathcal{F}_{src}^c$ as candidates for $\text{S}_1$. 
To obtain consolidated point features and queries, the pseudo queries $\mathcal{Q}_{ps}$ are further processed by the second encoder $\text{M}_{E_2}$ and refined through $\text{S}_2$, using randomly sampled input points $\mathcal{P}_{src}^f$ as augmentation.
The final queries $\mathcal{Q}$ are then aggregated using the max-pooled features $f_{g_1}$ and $f_{g_2}$ from the outputs $\mathcal{F}_1$ and $\mathcal{F}_2$ of the two encoders, together with the selected query points $\mathcal{P}_Q$.
This resampling and balancing strategy allows the model to adapt to different levels of input reliability: 
when input information is sparse or highly corrupted, 
the network relies more heavily on shape priors from the pseudo-queries; 
conversely, when the input is relatively clean or locally consistent, 
more input points are preserved as queries to retain geometric details.
Finally, the predicted points proxies $\mathcal{H}$ 
are generated by a GED $\text{M}_{D_2}$ from the selected queries $\mathcal{Q}$ and
the refined features $\mathcal{V}$:
\begin{equation}
    \mathcal{H} = \text{M}_{D_2}(\mathcal{Q},\mathcal{V}).
    \label{eq:dec}
\end{equation}

\subsection{Adaptive Query Generation}
\paragraph{Multi-Scale Feature Extraction}
To achieve better feature abstraction from the input point cloud, we adopt the EdgeConv head from DGCNN \cite{wang2019dynamic} hierarchically
and append light-weighted transformer encoder blocks to refine the features in each level of the transition down process. 
For level $l \in \{0,1,2\}$, given the input points 
%$\mathcal{P}_{\tilde l} = \{P_i\}_{i=1}^{N_l}$ with $P_i \in \mathbb{R}^3$ 
$\mathcal{P}^{\downarrow}_l = \{P_i\}_{i=1}^{N_l}$ with $P_i \in \mathbb{R}^3$ 
and the corresponding features 
%$\mathcal{F}_{\tilde l} = \{F_i\}_{i=1}^{N_l}$ with $F_i \in \mathbb{R}^{C_\tilde l}$,
$\mathcal{F}^{\downarrow}_l = \{F_i\}_{i=1}^{N_l}$ with $F_i \in \mathbb{R}^{C_l}$, 
the downsampled points 
%$\mathcal{P}_{\widetilde {l+1}}$ 
$\mathcal{P}^{\downarrow}_{l+1}$ 
and features 
%$\mathcal{F}_{\widetilde {l+1}}$ 
$\mathcal{F}^{\downarrow}_{l+1}$ 
in level ${l+1}$ can be obtained by:
\begin{equation}
    %\mathcal{P}_{\widetilde{l+1}}=\text{FPS}(\mathcal{P}_{\tilde l}, M_{\tilde{l}}),\quad \mathcal{F}_{\widetilde {l+1}}=\mathcal{E}(\text{EC}(\mathcal{F}_{\tilde l}, \mathcal{P}_{\tilde {l+1}})),
    \mathcal{P}^{\downarrow}_{l+1}=\text{FPS}(\mathcal{P}^{\downarrow}_l, M_l),\quad \mathcal{F}^{\downarrow}_{l+1} =\mathcal{E}(\text{EC}(\mathcal{F}^{\downarrow}_l, \mathcal{P}^{\downarrow}_{l+1})),
    \label{eq:down}
\end{equation}
where FPS is the farthest point sampling operation \cite{qi2017pointnet++} to downsample the points to $M_l$ points,
$\mathcal{E}$ denotes the transformer encoder blocks, and EC represents the EdgeConv operation indexed by the downsampled points.
The outputs from the last level form the coarse level point coordinates $\mathcal{P}_{src}^c$ and features $\mathcal{F}_{src}^c$ as shown in \cref{fig:model}.

\paragraph{Geo-Embedding Self-Attention}
\begin{figure}[t]
    \centering
    \includegraphics[width=0.99\linewidth]{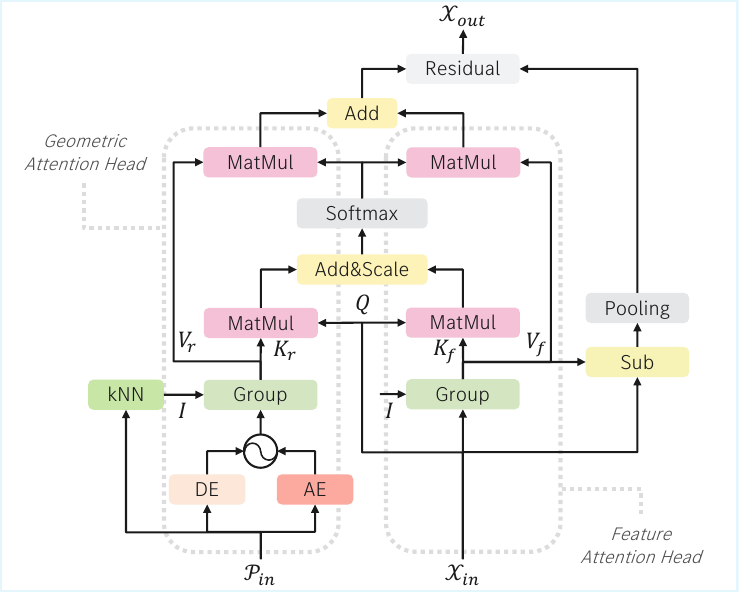}
    \caption{\textbf{Geometric-embedding self-attention block.}
    Geometric attention head uses distance embedding (DE) and angular embedding (AE) from input point coordinates 
    and regrouped as attention keys $K_r$ and values $V_r$. 
    For example, $\mathcal{P}_{in}$, $\mathcal{X}_{in}$ of the first self-attention block in $\text{M}_{E_1}$ correspond to $\mathcal{P}_{src}^c$ and $\mathcal{F}_{src}^c$, respectively.
    %Feature attention head gathers the input pointwise feature with the same $k$NN indices as key $k_f$ and value $v_f$. 
    %The input feature serves as query $q$ for both sides. 
    %Attention scores from two heads are added together and aggregated with residual input feature refined with EdgeConv layer.
    }
    \label{fig:attn}
    \vspace{-10pt}
\end{figure}
The downsampled points and their associated features are then processed by a series of transformer blocks to capture global geometric structures and enable geometric transformation.
We note that a positional encoding is crucial for the transformer to learn the spatial relationships among 3D points.
Prior works \cite{yu2021pointr,chen2023anchorformer,zhou2022seedformer,zhao2021point,li2023proxyformer} 
employ either absolute or relative encodings based on point coordinates or distance differences.

We propose a more comprehensive sparse geometric-embedding (SGE) strategy to encode transformation-invariant spatial relationships.
Given coarse-level points $\mathcal{P}_{in} = \{ {P_i}\} _{i = 1}^M$, the dense geometric structure embedding \cite{qin2023geotransformer,yu2023rotation} is defined as
\begin{equation}
    r_{i,j} = \mathbf{r}^{D}_{i,j}\mathbf{W}^{D} 
    + \max_{x} \left\{ \mathbf{r}^{A}_{i,j,x}\mathbf{W}^{A} \right\},
    \label{eq:dense_emb}
\end{equation}
where $\mathbf{r}^{D}_{i,j} \in \mathbb{R}^{C_e}$ encodes pair-wise distances and $\mathbf{r}^{A}_{i,j,x} \in \mathbb{R}^{C_e}$ encodes triplet-wise angles with $x$ as the index of a third neighbor of $P_i$, 
where $C_e$ is the internal dimension of the geometric embedding.
$\mathbf{W}^{D},\mathbf{W}^{A} \in \mathbb{R}^{ C_e\times C_e}$ are the corresponding projection matrices.
Directly using this dense embedding in self-attention is computationally expensive due to all pair- and triplet-wise interactions.
To improve efficiency, we adopt a sparse formulation by limiting the relations to the $k$-nearest neighbors for each point:
\begin{equation}
    \mathcal{R}_s = \left\{ r_{i,j} \;\middle|\; j \in \mathcal{N}_k(i), \; i=1,\dots,M \right\},
    \label{eq:sparse_emb}
\end{equation}
where $\mathcal{N}_k(i)$ denotes the set of indices of the $k$-nearest neighbors of point $P_i$.
Again, $r_{i,j}$ follows the definition from \cref{eq:dense_emb}.
This sparse representation not only retains the essential geometric relationships but also significantly reduces
the size of the embedding from $\mathcal{R} \in \mathbb{R}^{M \times M \times C_e}$ to
$\mathcal{R}_s \in \mathbb{R}^{M \times k \times C_e}$, where $k \ll M$.

The corresponding attention score of the attention head with SGE can be formulated as:
\begin{equation}
    \text{head} = \text{softmax}\!\left(\frac{Q (K_f+K_r)^{\top}}{\sqrt{d_h}}\right) (V_f+V_r),
    \label{eq:score_sge}
\end{equation}
where $Q$ is derived from the input features $\mathcal{X}_{in}$ feeding into $\text{M}_{E_{1,2}}$, $\text{M}_{D_{1,2}}$ in \cref{fig:model}. 
As shown in \cref{fig:attn}, the projected keys and values $K_f,V_f$ are regrouped via $k$NN indices, 
and combined with geometric embeddings $K_r=\mathcal{R}_s\mathbf{W}^{K_r}$ 
and extra geometric cue \cite{yu2023rotation} $V_r=\mathcal{R}_s\mathbf{W}^{V_r}$, 
all in $\mathbb{R}^{M \times k \times d_h}$.
Attention scores from two heads are added together and aggregated with residual input feature refined with an EdgeConv layer as final output $\mathcal{X}_{out}$.

\paragraph{Dynamic Query Selection}
Since the decoder output $\mathcal{H}$ in \cref{eq:dec} depends on the query embeddings $\mathcal{Q}$ 
and pseudo-queries $\mathcal{Q}_{ps}$ in \cref{eq:enc2}, 
the quality of these sets is critical to reconstruction performance. 
We introduce two Dynamic Query Selection (DQS) modules, $\text{S}_1$ and $\text{S}_2$, 
to adaptively filter out redundant or noisy candidate points. 
Because we mix a portion of input points as padding queries to preserve geometric fidelity, 
the DQS modules also balance the ratio of predicted and input points.

\begin{figure}[t]
    \centering
    \includegraphics[width=0.99\linewidth]{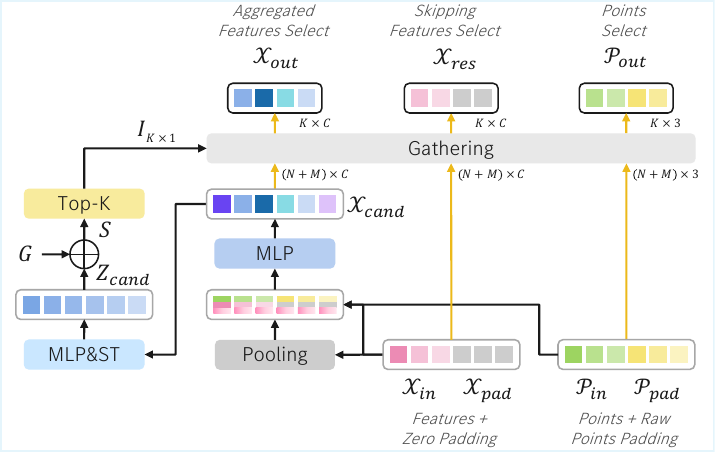}
    \caption{\textbf{Dynamic Query Selection (DQS) module.}
    Given the input coordinates $P_{in}$ and features $X_{in}$, 
    we expand the points with padding points $P_{pad}$ and zero-padded features $X_{pad}$ to a fixed size. 
    A Gumbel top-$k$ sampling strategy is then applied, 
    using aggregated scores to select the most representative points as output queries.}
    \label{fig:dqs}
    \vspace{-10pt}
\end{figure}

Unlike the learnable Gumbel-Top-$k$ subset sampling~\cite{xie2019subsets,kool2019stochastic}, 
we adopt a simple, non-learnable \emph{Perturb-and-Top-$k$} strategy, which yields better stability and performance. 
As shown in \cref{fig:dqs}, given the candidate features $\mathcal{X}_{cand}$ aggregated from the padded inputs, 
we compute standardized scores $\mathcal{Z}_{cand} = (\text{MLP}(\mathcal{X}_{cand}) - \mu)/\sigma$, 
perturb them with Gumbel noise $g_i = -\log(-\log(\text{Uniform}(0,1)))$, and obtain:
\begin{equation}
    s_i = z_i + \beta g_i, \quad i=1,\dots,N,
    \label{eq:gumbel_noise}
\end{equation}
where the noise scale $\beta$ is gradually annealed from $1.0$ to $0$ through training. 
We select the indices of the top-$k$ scores:
\begin{equation}
    I = \operatorname{Top}\text{-}k(\{s_i\}_{i=1}^{N}).
    \label{eq:topk}
\end{equation}

Compared to deterministic Top-$k$ selection, which may overfit by repeatedly choosing the same high-logit points, 
the Gumbel perturbation introduces controlled stochasticity that encourages exploration. Using a straight-through estimator \cite{bengio2013estimating}, gradients propagate through the selected feature paths while stochastic sampling allows different candidates to be explored across iterations.
This acts as a biased but low variance estimator and data-dependent regularizer, improving robustness and generalization. Further details of the gradient approximation is provided in the supplementary material.

\subsection{Coarse-to-Fine Expansion}
The output of the second transformer decoder are the high-dimensional point proxies 
$\mathcal{H} = \{ {H_i}\} _{i = 1}^M$ with $H_i \in \mathbb{R}^{C_H}$ containing the contextual shape information.
To unwrap the detailed geometry from these proxies in a higher resolution,
we feed the point proxies $\mathcal{H}$ as the input feature $\mathcal{H}^{\uparrow}_0$ to the hierarchical upsample transformer layers \cite{zhou2022seedformer} 
and utilize the query coordinates $\mathcal{P}_Q$ as seed and first level coordinates $\mathcal{P}^{\uparrow}_0$. For each level $l \in \{0,1,2\}$ of the transition up process, we have:
\begin{equation}
    \mathcal{P}^{\uparrow}_{l+1}, \mathcal{H}^{\uparrow}_{l+1}=\text{UpTrans}(\mathcal{P}_Q, \mathcal{P}^{\uparrow}_l, \mathcal{H}^{\uparrow}_l).
    \label{eq:up}
\end{equation}
We avoid folding-based upsampling methods \cite{yang2018foldingnet,yuan2018pcn,yu2021pointr} and MLP-based projections \cite{Yu2023AdaPoinTrDP,wei2025pcdreamer}, 
which generate patches independently and ignore local geometric relationships.
In contrast, coarse-to-fine upsampling \cite{xiang2021snowflakenet,zhou2022seedformer} 
with geometric context refinement in each level provides a more fine-grained and 
stable manner to recover the detailed geometry.

\subsection{Network Optimization}
Since our model generates point clouds in a coarse-to-fine manner, each prediction level $\mathcal{P}^{\uparrow}_i$ is supervised by the ground-truth point cloud $\mathcal{G}$ using the L1 Chamfer Distance (CD$_{\ell1}$) \cite{fan2017point}.
To ensure balanced supervision, $\mathcal{G}$ is downsampled via FPS to match the number of points in $\mathcal{P}^{\uparrow}_i$, preventing over-penalization of the backward CD term when $\mathcal{P}^{\uparrow}_i$ is sparser.
This strategy stabilizes training at coarse stages by avoiding penalties for unrecoverable fine details.
The multi-level reconstruction loss $\mathcal{L}_{rec}$ is defined as:
\begin{equation}
    \mathcal{L}_{rec} = \sum_{i=1}^{L} \text{CD}_{\ell1}(\mathcal{P}^{\uparrow}_i, \text{FPS}(\mathcal{G}, |\mathcal{P}^{\uparrow}_i|)).
    \label{eq:loss_rec}
\end{equation}
The intermediate pseudo-query points $\mathcal{P}_{pq}$ are also constrained by the ground truth $\mathcal{G}$ 
with a separate loss $\mathcal{L}_{pq}$:
\begin{equation}
    \mathcal{L}_{pq} = \text{CD}_{\ell1}(\mathcal{P}_{pq}, \text{FPS}(\mathcal{G}, |\mathcal{P}_{pq}|)).
    \label{eq:loss_ps}
\end{equation}
The overall loss function is the sum of the above two losses:
\begin{equation}
    \mathcal{L} = \mathcal{L}_{rec} + \mathcal{L}_{pq}.
    \label{eq:loss}
\end{equation}
%where $\lambda$ is a balancing weight to control the trade-off between intermediate pseudo-query supervision 
%and final reconstruction quality.

\section{Experiments}
\label{sec:exp}
\begin{table*}[t]
  \caption{\textbf{Results of our method and state-of-the-art methods on ShapeNet-55/34.} 
  We report the results of all 55 categories, 34 seen categories and 21 unseen categories in three difficulty degrees. 
  We use CD-S, CD-M and CD-H to represent the CD$_{\ell2}$ (multiplied by 1000) results under the Simple, Moderate and Hard settings. 
  We also provide average results among different settings under the CD$_{\ell2}$ and F-Score@1\% metric.}
  \centering
  \footnotesize
  \setlength{\tabcolsep}{3.4pt} 
  \begin{tabular}{l 
  ccc|>{\columncolor{gray!10}}c>{\columncolor{gray!10}}c 
  ccc|>{\columncolor{gray!10}}c>{\columncolor{gray!10}}c 
  ccc|>{\columncolor{gray!10}}c>{\columncolor{gray!10}}c}
  \toprule
  \multirow{3}{*}{Method} & \multicolumn{5}{c}{\textbf{All 55 categories}} & \multicolumn{5}{c}{\textbf{34 seen categories}} & \multicolumn{5}{c}{\textbf{21 unseen categories}} \\
  \cmidrule(rl){2-6}\cmidrule(rl){7-11}\cmidrule(rl){12-16}
    & CD-S & CD-M & CD-H & CD$_{\ell2}$ ($\downarrow$) & F1 ($\uparrow$) 
    & CD-S & CD-M & CD-H & CD$_{\ell2}$ ($\downarrow$) & F1 ($\uparrow$)
    & CD-S & CD-M & CD-H & CD$_{\ell2}$ ($\downarrow$) & F1 ($\uparrow$) \\
  \midrule
  FoldingNet\cite{yang2018foldingnet} & 2.67 & 2.66 & 4.05 & 3.12 & 0.082
  & 1.86 & 1.81 & 3.38 & 2.35 & 0.139 & 2.76 & 2.74 & 5.36 & 3.62 & 0.095 \\
  PCN\cite{yuan2018pcn} & 1.94 & 1.96 & 4.08 & 2.66 & 0.133
  & 1.87 & 1.81 & 2.97 & 2.22 & 0.154 & 3.17 & 3.08 & 5.29 & 3.85 & 0.101 \\
  GRNet\cite{xie2020grnet} & 1.35 & 1.71 & 2.85 & 1.97 & 0.238
  & 1.26 & 1.39 & 2.57 & 1.74 & 0.251 & 1.85 & 2.25 & 4.87 & 2.99 & 0.216 \\
  PoinTr\cite{yu2021pointr} & 0.58 & 0.88 & 1.79 & 1.09 & 0.464
  & 0.76 & 1.05 & 1.88 & 1.23 & 0.421 & 1.04 & 1.67 & 3.44 & 2.05 & 0.384 \\
  SnowflakeNet \cite{xiang2021snowflakenet} & 0.70 & 1.06 & 1.96 & 1.24 & 0.398
  & 0.60 & 0.86 & 1.50 & 0.99 & 0.422 & 0.88 & 1.46 & 2.92 & 1.75 & 0.388 \\
  SeedFormer\cite{zhou2022seedformer} & 0.50 & 0.77 & 1.49 & 0.92 & 0.472
  & 0.48 & 0.70 & 1.30 & 0.83 & 0.452 & 0.61 & 1.07 & 2.35 & 1.34 & 0.402 \\
  AdaPoinTr \cite{Yu2023AdaPoinTrDP} & 0.49 & 0.69 & 1.24 & 0.81 & 0.503
  & 0.48 & 0.63 & 1.07 & 0.73 & 0.469 & 0.61 & 0.96 & 2.11 & 1.23 & 0.416 \\
  ProxyFormer \cite{li2023proxyformer} & 0.49 & 0.75 & 1.55 & 0.93 & 0.483
  & 0.44 & 0.67 & 1.33 & 0.81 & 0.466 & 0.60 & 1.13 & 2.54 & 1.42 & 0.415 \\
  T-CorresNet \cite{duan2024t} & 0.50 & 0.68 & 1.23 & 0.80 & 0.485
  & 0.48 & 0.63 & 1.09 & 0.73 & 0.462 & 0.58 & 0.91 & 1.97 & 1.15 & 0.409 \\
  AnchorFormer\cite{chen2023anchorformer} & 0.41 & 0.61 & 1.26 & 0.76 & 0.558
  & 0.41 & 0.57 & 1.12 & 0.70 & 0.564 & 0.52 & 0.90 & 2.16 & 1.19 & 0.535 \\
  \midrule
  \textbf{PQDT} & \textbf{0.34} & \textbf{0.55} & \textbf{1.13} & \textbf{0.68} & \textbf{0.570}
  & \textbf{0.33} & \textbf{0.50} & \textbf{0.99} & \textbf{0.60} & \textbf{0.576} & \textbf{0.40} 
  & \textbf{0.75} & \textbf{1.81} & \textbf{0.99} & \textbf{0.548} \\
  \bottomrule
  \end{tabular}
  \label{tab:shapenet5534}
  \vspace{-10pt}
\end{table*}

First, we evaluate our proposed PQDT framework on the point cloud completion benchmark 
\textbf{ShapeNet-55/34} \cite{yu2021pointr,wu20153d}. Furthermore, we introduce new datasets called 
\textbf{ShapeNet-Deform} and \textbf{ShapeNetCar-Occ} to validate the robustness of our model under different types 
of noise, distortion or difference input quality. To show the local geometry restoration ability of our model,
we also perform experiments on the patchified freeform surface (\textbf{PFS}) dataset.
We present the results of our model compared with state-of-the-art methods, and ablation studies further validate the effectiveness of the proposed modules.

\subsection{Datasets and Benchmarks}
\paragraph{Completion-only Benchmarks}
For point cloud restoration, we adopt the widely used ShapeNet-55/34 dataset, instead of PCN \cite{yuan2018pcn}, 
since it offers more diverse object categories and more challenging incomplete shapes. 
We follow the standard experimental setup from prior works \cite{yu2021pointr,zhou2022seedformer,chen2023anchorformer}, 
evaluating three levels of incompleteness. 

\begin{figure}[t]
    \centering
    \includegraphics[width=0.99\linewidth]{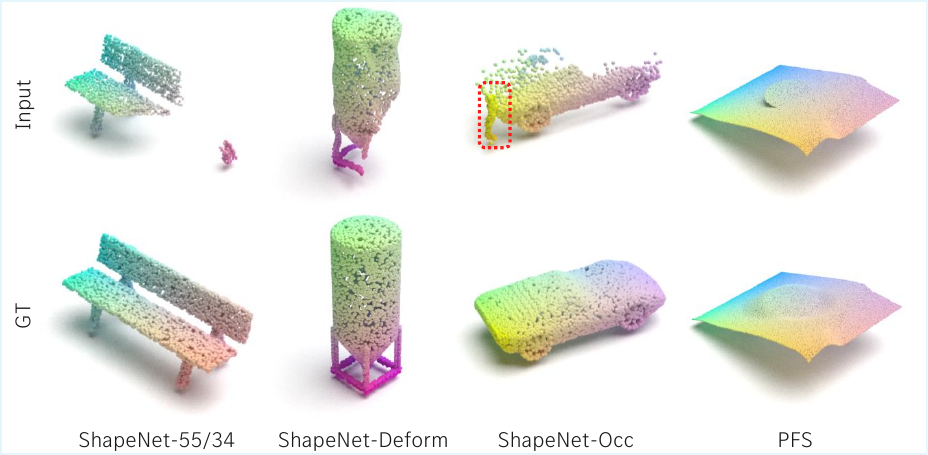}
    \caption{Visualization of datasets used for evaluation.
%    ShapeNet-Deform is generated from ShapeNet55/34 by applying Gabor noise as geometric distortion, 
%    and both datasets use cropping to create incomplete point clouds. 
%    ShapeNetCar-Occ is an occluded augmentation of the car category, where partial point clouds are generated via raycasting. 
%    PFS contains cropped freeform surfaces of Body-in-White data before and after restoration.
}
    \label{fig:datasets}
    \vspace{-10pt}
  \end{figure}

\paragraph{Restoration Benchmarks}
To evaluate robustness under geometric distortions, we introduce the ShapeNet-Deform dataset by applying Gabor noise \cite{lagae2009gabor} to ShapeNet shapes, producing smooth and anisotropic deformations.
Unlike Gaussian noise, Gabor noise generates more natural geometric variations via convolution with Gabor kernels.
We vary kernel parameters (frequency, scale, and Gaussian width) to create three deformation levels.
To assess robustness to occlusion and sensor noise, we construct ShapeNetCar-Occ, 
a dataset of car models with synthetic occluders (e.g., pedestrians, signs) and LiDAR-style raycasting \cite{levoy1988display,zhou2018open3d}.
Each of the 3,419 car instances is rendered from 32 random viewpoints under three noise and occlusion settings, 
with additional Gaussian noise applied for realism.
\cref{fig:datasets} shows examples.

\paragraph{PFS Dataset}
Freeform surfaces are widely used in CAD and computer graphics to represent complex geometries. 
A common application is surface restoration from 3D scans as part of the reverse engineering process. 
In practice, this often involves repairing locally damaged regions or refining reference geometries (e.g., planar or curved surfaces) 
with user-guided local details. Unlike typical shape completion or deformation tasks, 
this type of restoration requires a combination of completion and deformation while preserving global consistency, 
since only specific local areas are intended to be modified. 
To support research in this challenging setting, we construct the Patchified Freeform Surface (PFS) dataset, 
derived from Body-in-White components in the automotive industry. 
The PFS dataset contains 880 freeform surface patches, split into 800 for training and 80 for testing. 
Each sample includes the geometry before and after restoration, along with a guidance surface, as illustrated in \cref{fig:datasets}. 
Further dataset statistics and experimental details are provided in the supplementary material.

\paragraph{Evaluation Metrics}
We follow previous works \cite{xie2020grnet,yuan2018pcn,zhou2022seedformer} 
for the point cloud completion task to evaluate our extended benchmarks with corresponding datasets.
We adopt the CD$_{\ell2}$ and  the F-Score \cite{tatarchenko2019single} as evaluation metrics.
\begin{figure*}[t]
    \centering
    \includegraphics[width=0.99\linewidth]{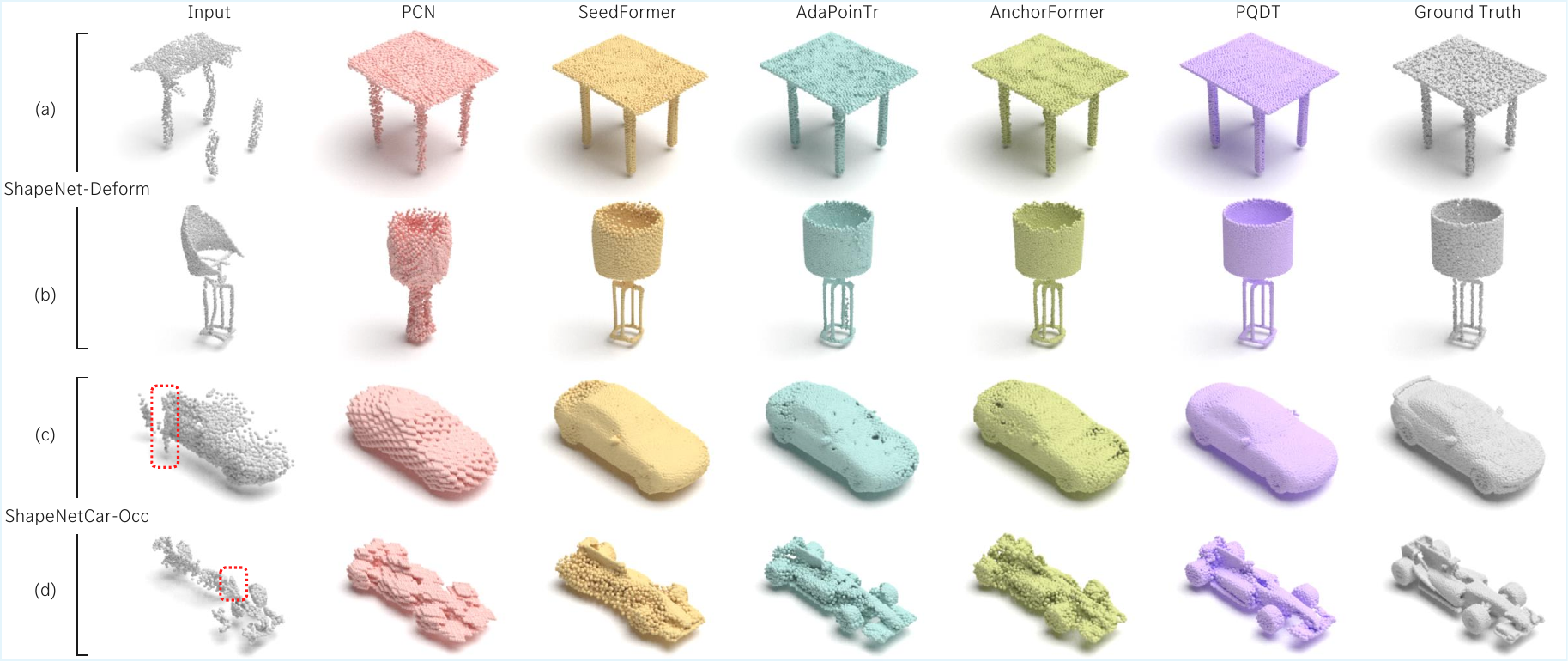}
    \caption{\textbf{Qualitative evaluation on ShapeNet-Deform and ShapeNetCar-Occ.}
    Red boxes indicate occluding objects.
    }
    \label{fig:vis_2d2o}
\end{figure*}
\begin{table*}[h]
  \caption{\textbf{Results of our method and state-of-the-art methods on ShapeNet-Deform, ShapeNetCar-Occ and PFS dataset.} 
  We report the CD$_{\ell2}$ (multiplied by 1000) under three difficulty setups of ShapeNet-Deform, 
  and the average CD$_{\ell2}$ and F-Score@1\% of all three datasets. Additional F-Score@0.5\% are reported for PFS dataset.
  }
  \centering
  \footnotesize
  \setlength{\tabcolsep}{9.1pt} 
  \begin{tabular}{l ccc|cc cc ccc}
  \toprule
  \multirow{3}{*}{Method} & \multicolumn{5}{c}{\textbf{ShapeNet-Deform}} & \multicolumn{2}{c}{\textbf{ShapeNetCar-Occ}} & \multicolumn{3}{c}{\textbf{PFS dataset}}\\
  \cmidrule(rl){2-6}\cmidrule(rl){7-8}\cmidrule(lr){9-11}
    & CD-S & CD-M & CD-H & CD$_{\ell2}$ ($\downarrow$) & F1 ($\uparrow$) 
    & CD$_{\ell2}$ ($\downarrow$) & F1 ($\uparrow$) 
    & CD$_{\ell2}$ ($\downarrow$) & F1 ($\uparrow$) & F0.5 ($\uparrow$) \\
  \midrule
    PCN \cite{yuan2018pcn} & 1.81 & 2.10 & 3.08 & 2.33 & 0.162 
    & 1.42 & 0.125 
    & 1.60 & 0.233 & 0.040 \\
    PoinTr \cite{yu2021pointr} & 0.85 & 1.81 & 2.10 & 1.59 & 0.204 
    & 1.07 & 0.191 
    & 0.92 & 0.376 & 0.071 \\
    SnowflakeNet \cite{xiang2021snowflakenet} & 0.96 & 1.26 & 2.15 & 1.46 & 0.229 
    & 0.94 & 0.220 
    & 1.04 & 0.656 & 0.222 \\
    SeedFormer \cite{zhou2022seedformer} & 0.82 & 1.16 & 2.21 & 1.40 & 0.235 
    & 0.97 & 0.231 
    & 0.88 & 0.619 & 0.183 \\
    AdaPoinTr \cite{Yu2023AdaPoinTrDP} & 0.67 & 0.89 & \textbf{1.53} & 1.03 & 0.288 
    & 0.85 & 0.251 
    & 0.69 & 0.603 & 0.187 \\
    AnchorFormer \cite{chen2023anchorformer} & 0.75 & 0.98 & 1.68 & 1.14 & 0.270 
    & 0.93 & 0.225 
    & 1.17 & 0.406 & 0.082 \\
  \midrule
    \textbf{PQDT} & \textbf{0.63} & \textbf{0.85} & 1.54 & \textbf{1.01} & \textbf{0.290} 
    & \textbf{0.82} & \textbf{0.261} 
    & \textbf{0.16} & \textbf{0.834} & \textbf{0.311} \\
  \bottomrule
  \end{tabular}
  \label{tab:deform_occ_pfs}
  \vspace{-10pt}
\end{table*}
\subsection{Evaluation Results}
We compare PQDT with several 3D-only state-of-the-art methods \cite{yang2018foldingnet,yuan2018pcn,xie2020grnet,yu2021pointr,xiang2021snowflakenet,zhou2022seedformer,Yu2023AdaPoinTrDP,li2023proxyformer,duan2024t,chen2023anchorformer}. 
We report both quantitative and qualitative results on the introduced datasets. 
For ShapeNet-55/34, we adopt results reported in the literature, while all baselines are re-implemented for our newly introduced datasets to ensure fair comparison. 
ProxyFormer \cite{li2023proxyformer} and T-CorresNet \cite{duan2024t} are excluded from the experiments on the latter three datasets, as their official implementations are not fully available.
Note that we omit the recent diffusion-based methods SuperPC \cite{du2025superpc} and PCDreamer \cite{wei2025pcdreamer}, since these require additional 2D image inputs alongside point clouds. 

\paragraph{Results on ShapeNet-55/34}
As summarized in Table \ref{tab:shapenet5534}, PQDT achieves new state-of-the-art results across all evaluation metrics on ShapeNet-55.
We report the average CD$_{\ell2}$ and F1-Scores over all categories, as well as CD-S, CD-M, 
and CD-H metrics corresponding to three levels of incompleteness. 
%Following the evaluation protocol of prior works \cite{yu2021pointr,chen2023anchorformer}, 
%we also include CD$_{\ell2}$ results for categories with the most and least training samples (details in SuppXXX) to validate the model’s robustness under limited data.
Across all settings, PQDT consistently outperforms the previous best method, AnchorFormer, 
with a notable improvement of 0.13 in the hard cases (CD-H).
Overall, PQDT attains an average CD$_{\ell2}$ of 0.68 and F1-Score of 0.57 across all difficulty levels, 
demonstrating superior completion accuracy and surface consistency.
The table further shows the adaptability of PQDT on ShapeNet-34. 
The model consistently performs best on both seen and unseen categories with noticeable advantages in CD and F-Score. 
The results confirm PQDT’s robust generalization and its ability to preserve fine-grained structure across diverse object types.

\paragraph{Results on ShapeNet-Deform}
Recovering 3D shapes from distorted and incomplete partial inputs remains a highly challenging task. 
To assess the robustness of our method under such conditions, we evaluate PQDT on the ShapeNet-Deform dataset. 
As shown in \cref{tab:deform_occ_pfs}, PQDT achieves consistently superior performance across nearly all difficulty levels, 
surpassing strong baselines such as SeedFormer, AdaPoinTr and AnchorFormer. 
In particular, PQDT attains the lowest CD$_{\ell2}$ of 1.01 and highest F1-Score of 0.29, demonstrating its high reconstruction accuracy and robustness to deformation. 
These results clearly showcase PQDT’s ability to capture complex geometric variations and generate uniformly distributed, 
high-fidelity point clouds, reflecting strong generalization to challenging non-rigid scenarios.

\paragraph{Results on ShapeNetCar-Occ}
The ShapeNetCar-Occ dataset presents severe occlusion scenarios closer to real-world conditions, 
including inter-object occlusion from surrounding objects and self-occlusion caused by the object’s own geometry.
% making accurate completion particularly challenging. 
Our model achieves the best average CD$_{\ell2}$ of 0.82 and F1-Score of 0.261. 
This indicates that PQDT can effectively infer missing geometry and remains robust against local occlusion noise and global Gaussian noise. 
These results further validate the effectiveness of our query selection strategy, 
which leverages object-relevant information to reconstruct complete and coherent 3D shapes under challenging occlusion conditions.

\begin{figure}[t]
    \centering
    \includegraphics[width=0.99\linewidth]{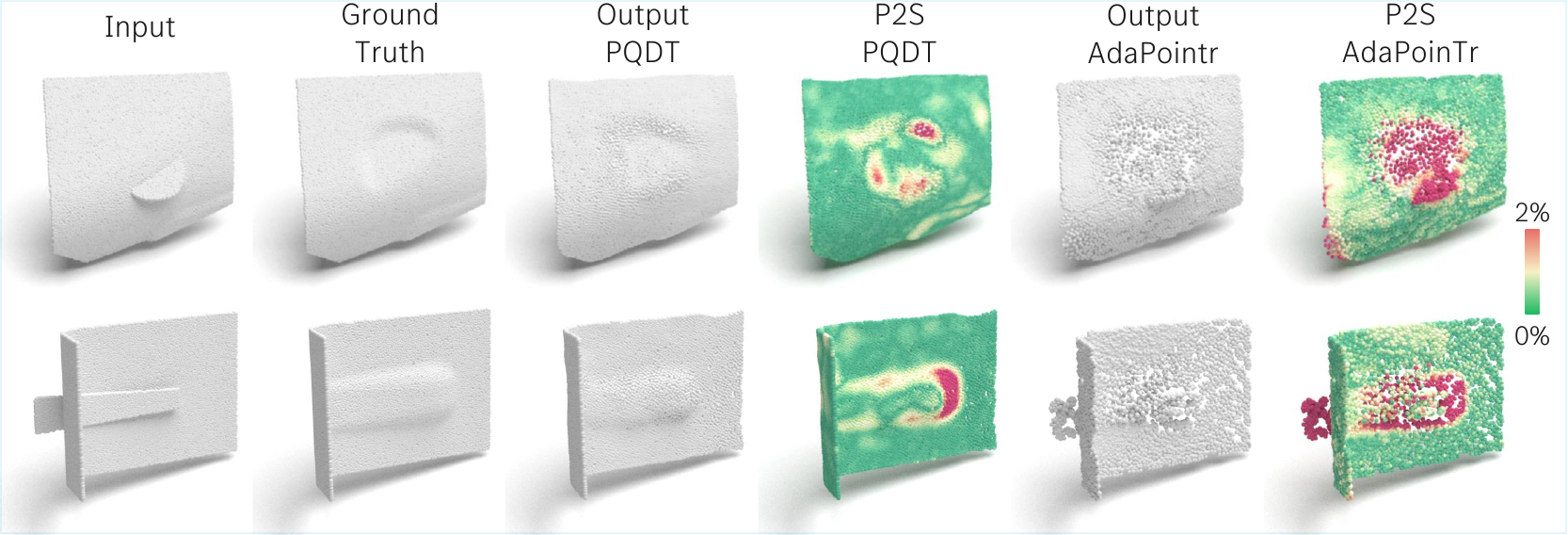}
    \caption{\textbf{Qualitative evaluation on PFS.}
    The color represent the Point-to-Surface distance (P2S) between generated point clouds and ground truth mesh, the maximum (red) is clamped with 2\% of the radius of the object’s bounding sphere.}
    \label{fig:vis_pfs}
    \vspace{-10pt}
\end{figure}

\paragraph{Results on PFS}
The PFS dataset is more representative of real-world industrial applications, 
emphasizing challenges in handling geometric deformations of freeform surfaces while maintaining input consistency. 
As shown in \cref{tab:deform_occ_pfs}, PQDT achieves the best overall performance, surpassing all competing methods by a large margin. 
Specifically, PQDT reduces CD$_{\ell2}$ by 76.8\% compared to the second-best method, 
while improving F1- and F0.5-Scores by 27.1\% and 40.1\%, respectively. 
From \cref{fig:vis_pfs} we can also see the qualitative results of our model compared with best-performed baseline AdaPointr. 
Our model shows the superior capability in producing consistent, deformation-aware, 
and high-fidelity point cloud reconstructions, demonstrating its potential for real-world industrial deployment.

\subsection{Model Analysis and Ablation Study}
\paragraph{Model Design}
To evaluate the effectiveness of our model design, we conduct an ablation study summarized in \cref{tab:ab_main}.
Starting from a vanilla Transformer for point clouds as the baseline (Model A), we progressively integrate our proposed components.
Model B introduces advanced query generation, including both the auxiliary pseudo-query and dynamic query selection modules.
This design yields an improvement in CD by 0.17 and a notable F1-Score gain of 0.012, demonstrating the effectiveness of our dual-stage transformer structure.
Model C further replaces the standard linear projection from global features with a query-based DETR-like decoder for seed proxy prediction, leading to consistent improvements in CD and confirming the stability of decoding seed coordinates from spherical priors.
Finally, our complete model PQDT incorporates the geometric-embedding (GE) in place of raw coordinate-based positional encoding, achieving the best overall performance.
These results verify that each proposed component contributes meaningfully to the final effectiveness of PQDT.
Detailed analyses for each design aspect are provided in the supplementary material.
\begin{table}[t]
  \caption{\textbf{Ablation study on ShapeNetCar-Occ.}
  We report the results with different model designs including query generation (Query), seed prediction (Seed) and positional encoding (PE).}
  \centering
  \footnotesize
  \setlength{\tabcolsep}{7.8pt}
  \begin{tabular}{c|ccc|cc}
  \toprule
  model & Query & Seed & PE & CD$_{\ell2}$ ($\downarrow$) & F1 ($\uparrow$) \\
  \midrule
  A & \checkmark & Linear Proj. & \checkmark & 0.857 & 0.243 \\
  B & Pseudo     & Linear Proj. & \checkmark & 0.840 & 0.255 \\
  C & Pseuso     & Query Dec.   & \checkmark & 0.827 & 0.258 \\
  PQDT & Pseudo  & Query Dec.   & GE         & 0.818 & 0.261 \\
  \bottomrule
  \end{tabular}
  \label{tab:ab_main}
  % \vspace{-10pt}
\end{table}

\begin{figure}[t]
    \centering
    \includegraphics[width=0.99\linewidth]{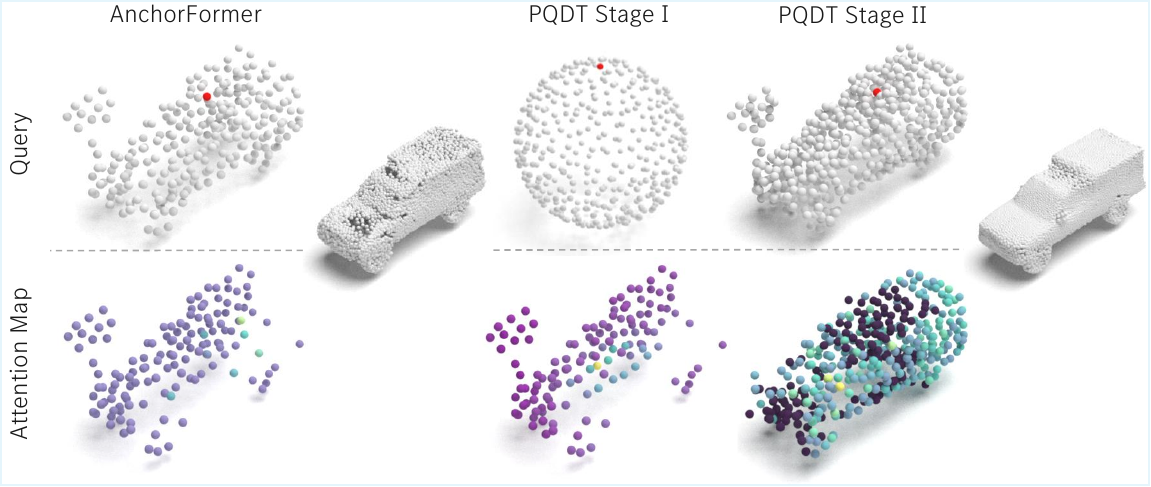}
    \caption{\textbf{Attention maps from selected queries (red dot) visualized on key points (bottom row).}
    Stage I attention maps show coarse pseudo-query exploration over the encoded latent structure. 
    Stage II maps demonstrate localized, coherent attention aligned with the underlying surface, indicating effective query refinement. The color reflects the attention score after the Softmax operation in the last block of the decoder.
    }
    \label{fig:attn_map}
    \vspace{-10pt}
\end{figure}

\paragraph{Attention Visualization}
To gain deeper insight into the behavior of our model relative to typical transformer-based architectures, we compare the query attention distributions with those of AnchorFormer, as shown in \cref{fig:attn_map}.
Each red point denotes a selected query, and the color intensity on the key points indicates the corresponding attention weights. 
As a representative state-of-the-art transformer baseline, AnchorFormer exhibits diffuse and weakly localized attention, suggesting less discriminative geometric reasoning. 
In contrast, PQDT Stage I produces pseudo-queries that roughly capture global shape context, 
while PQDT Stage II refines these queries into spatially coherent, part-aware representations. 
The attention in Stage II is sharply concentrated on geometrically relevant regions, 
demonstrating that the progressive query refinement mechanism effectively enhances local feature aggregation and geometric understanding.

\section{Conclusion}
\label{sec:conclusion}
We introduced a transformer-based framework for general point cloud restoration with a dual-stage pseudo-query design. Our point-only backbone effectively balances input observations and learned shape priors, achieving state-of-the-art performance across mixed restoration tasks involving completion, deformation, and denoising. The proposed model also generalizes well to newly introduced challenging scenarios, showing strong potential as a unified backbone for broader 3D perception tasks. 

\clearpage
\setcounter{page}{1}
\maketitlesupplementary
\appendix

In this supplementary material, we provide additional content to complement the main paper. 
\cref{sec:sup_impl} presents the full implementation details of PQDT, including training setups, network components.
\cref{sec:sup_metrics} gives the definition of the evaluation metrics.
\cref{sec:sup_data} describes the datasets used in our experiments and the pre-processing strategies applied. 
\cref{sec:sup_ab} reports extended ablation studies that analyze the contribution of each module in our design. 
\cref{sec:sup_eval} provides additional quantitative evaluation results across multiple benchmarks. 
Finally, \cref{sec:sup_sample} includes more qualitative visualizations to further illustrate the effectiveness of our approach.

\section{Implementation Details}
\label{sec:sup_impl}
\paragraph{Training Setups}
We implement PQDT using PyTorch \cite{paszke2019pytorch} and train the network with the Adam optimizer \cite{kingma2014adam}. The initial learning rate is set to 0.0001. We adopt a cosine annealing learning rate scheduler \cite{LoshchilovH17} with a minimum learning rate of 0.00001, along with a linear warm-up from 0.00001 to 0.0001 over the first 10 epochs. For the ShapeNet-55/34 and ShapeNet-Deform datasets, we train the model for 300 epochs with a batch size of 32. For smaller-scale datasets, ShapeNetCar-Occ and PFS, the batch sizes are set to 16 and 8, respectively, and the models are trained for 200 epochs. For all ShapeNet-family datasets, the network takes 2,048 points as input and predicts 8,192 points as output. On the PFS dataset, we instead sample 8,192 points from the geometry as inputs and predict 8,192 points as output. We train the model with a single Nvidia A100 80GB GPU, and all the models are tested with the batch size of 1 and same random seed.

\paragraph{Feature Extraction}
We employ a modified DGCNN \cite{wang2019dynamic} to extract features from the input point cloud. The network expands feature dimensionality while progressively downsampling points, following the architecture:

$\text{Linear}(C_{in}=3, C_{out}=8)
\rightarrow \text{EC}(C_{in}=8, C_{out}=64, K=16, M=512)
\rightarrow \mathcal{E}(C=64, H=1, L=2)
\rightarrow \text{EC}(C_{in}=64, C_{out}=256, K=16, M=128)
\rightarrow \mathcal{E}(C=256, H=4, L=2)$,\\
where $C_{\text{in}}, C_{\text{out}}$ denote input and output feature channels; EC is an EdgeConv block with $K$ nearest neighbors and farthest point sampling size $M$; and $\mathcal{E}$ is a lightweight transformer encoder with feature dimension $C$, number of heads $H$, and number of layers $L$.  
The resulting coarse level features are:
$\mathcal{F}^{c}_{src} \in \mathbb{R}^{128 \times 256}, 
\mathcal{P}^{c}_{src} \in \mathbb{R}^{128 \times 3}$.

\paragraph{Transformer}
As illustrated in \cref{fig:model}, the dual transformer module, comprising geometric-embedding transformer encoders and decoders (GEE/GED), denoted $\text{M}_{E_{1,2}}$ and $\text{M}_{D_{1,2}}$, processes the coarse features and coordinates as follows:

\textit{First-Stage Encoding}: 
$\text{M}_{E_1}(C=384, H=6, L=4)
\rightarrow \mathcal{F}_1
\rightarrow \text{MaxPool}
\rightarrow f_{g_1}$.

\textit{First-Stage Decoding}: 
$\text{Agg}(\text{FPS}(\mathcal{P}_{sph}(R=0.8), M=384), f_{g_1})
\rightarrow \text{MLP}(C_{in}=3+1024, C_{out}=384)
\rightarrow \text{M}_{D_1}(C=384, H=6, L=4, \mathcal{V}=\mathcal{F}_1)
\rightarrow \mathcal{F}_{pq'}
\rightarrow \text{MLP}(C_{in}=384, C_{out}=3)
\rightarrow \mathcal{P}_{pq'}$.
$\text{S}_1(N_{in}=384, N_{pad}=128, N_{out}=384,\ 
\mathcal{X}_{in}=\mathcal{F}_{pq'},\ 
\mathcal{P}_{in}=\mathcal{P}_{pq'})
\rightarrow 
\mathcal{F}_{pq},\ \mathcal{P}_{pq}$.

\textit{Second-Stage Encoding}: 
$\text{M}_{E_2}(C=384, H=6, L=4)
\rightarrow \mathcal{F}_2
\rightarrow \text{MaxPool}
\rightarrow f_{g_2}$.
$\text{S}_2(N_{in}=384, N_{pad}=256, N_{out}=512,\ 
\mathcal{X}_{in}=\mathcal{F}_2,\ 
\mathcal{P}_{in}=\mathcal{P}_{pq})
\rightarrow \mathcal{V},\ \mathcal{P}_Q$.

\textit{Second-Stage Decoding}: 
$\text{Agg}(\mathcal{P}_Q, f_{g_1}, f_{g_2})
\rightarrow \text{MLP}(C_{\in}=3+1024+1024, C_{out}=384)
\rightarrow \mathcal{Q}
\rightarrow 
\text{M}_{D_2}(C=384, H=6, L=8, \mathcal{V}=\mathcal{V})
\rightarrow \text{MaxPool}
\rightarrow f_{g_q}$.

\textit{Output Aggregation}: 
$\text{Agg}(\mathcal{P}_Q, \mathcal{Q}, f_{g_q})
\rightarrow \text{MLP}(C_{in}=3+384+1024, C_{out}=384)
\rightarrow \mathcal{H}$.

Here, Agg denotes feature alignment and concatenation.  
$\text{S}_{1,2}$ denote dynamic query selection modules with input size $N_{in}$, padding size $N_{pad}$, and output size $N_{out}$.  
$\mathcal{P}_{sph}$ is a point set sampled from a sphere of radius $R$.  
Features preceding max pooling are projected to 1024 channels.  
The final coarse prediction outputs are:
$\mathcal{F}^{c}_{pred}=\mathcal{H} \in \mathbb{R}^{512 \times 384}, 
\mathcal{P}^{c}_{pred}=\mathcal{P}_Q \in \mathbb{R}^{512 \times 3}$.

\paragraph{Geometric-Embedding}
We follow the calculation of distance and angular embedding from \cite{qin2023geotransformer} to form our sparse geometric-embedding (SGE):

\textit{Pair-wise Distance Embedding}: By applying the sinusoidal encoding \cite{vaswani2017attention} to the Euclidean distance 
$\rho_{i,j} = \lVert P_i - P_j \rVert_2$ between points $P_i$ and $P_j$, the distance embedding is defined as:
\begin{equation}
\label{eq:dist-embed}
\left\{
\begin{aligned}
r^{D}_{i,j,2k} &= \sin\!\left(\frac{\rho_{i,j}/\sigma_d}{10000^{2k/C_e}}\right), \\
r^{D}_{i,j,2k+1} &= \cos\!\left(\frac{\rho_{i,j}/\sigma_d}{10000^{2k/C_e}}\right),
\end{aligned}
\right.
\end{equation}
where $\sigma_d=0.2$ is the temperature which is used to tune the sensitivity on distance variation and we use same value as in the original work. $C_e$ is the channel size of embedding.

\textit{Triplet-wise Angular Embedding}: The angular embedding for a point $P_i$ is computed using its $k$ nearest neighbors $\mathcal{K}_i$. 
For each neighbor $P_x \in \mathcal{K}_i$, we evaluate the angle $\alpha^x_{i,j}=\angle(\Delta_{x,i},\Delta_{j,i})$, where $\Delta_{i,j}=P_i-P_j$. Applying a sinusoidal positional encoding to $\alpha^{x}_{i,j}$, we obtain the angular embedding:
\begin{equation}
\label{eq:angle-embed}
\left\{
\begin{aligned}
r^{A}_{i,j,x,2l} &= \sin\!\left(\frac{\alpha^{x}_{i,j}/\sigma_a}{10000^{2l/C_e}}\right), \\
r^{A}_{i,j,x,2l+1} &= \cos\!\left(\frac{\alpha^{x}_{i,j}/\sigma_a}{10000^{2l/C_e}}\right),
\end{aligned}
\right.
\end{equation}
where $\sigma_\alpha=15$ is the temperature which control the sensitivity on angular variations. Then we assemble these two types of embeddings can compute SGE using \cref{eq:dense_emb,eq:sparse_emb}.

\paragraph{Spherical Initialization}
We adopt a DETR-like decoder \cite{carion2020end} initialized with a set of learnable spherical queries (see \cref{fig:attn_map}). Specifically, we empirically sample 384 queries on the surface of a fixed sphere with radius $0.8$, while all input point clouds are normalized to lie within the unit sphere. This design bounds the displacement between the initial queries and the target geometry, enabling the network to deform the spherical points toward missing regions via cross-attention. Extremely large missing regions may complicate global scale estimation and introduce potential bias, but such cases are uncommon in standard benchmarks. Some queries may not correspond to meaningful regions of the final geometry; however, the attention mechanism naturally suppresses such queries during decoding. The resulting redundancy increases flexibility in modeling diverse geometric structures without introducing noticeable computational overhead. Overall, the spherical initialization provides a stable geometric prior that facilitates robust and consistent shape completion.

\paragraph{Gradient approximation for Dynamic Query Selection}
In \cref{sec:method}, the query selection step involves Gumbel-perturbed Top-$k$ sampling followed by hard indexing, which makes the operation non-differentiable. Ideally, training would minimize the expected reconstruction loss over the stochastic selection process:
\begin{equation}
\mathcal{L}(\theta) 
= \mathbb{E}_{S \sim p_\theta(S)} 
\left[ \ell\big(f_\theta(X, S)\big) \right],
\end{equation}
where $X$ denotes the point features, $S$ denotes the selected candidate set, $p_\theta(S)$ is the selection distribution induced by the scoring network, $f_\theta$ denotes the downstream decoding function, and $\ell$ is the reconstruction loss. In principle, unbiased gradients of this objective require score-function estimators (e.g., REINFORCE \cite{williams1992simple}), which propagate gradients through the sampling probabilities but typically exhibit high variance and unstable optimization. Instead, we adopt a straight-through estimator that ignores gradients through the sampling distribution and backpropagates only through the selected feature paths~\cite{bengio2013estimating}. While this yields a biased gradient estimator, it significantly reduces variance and leads to stable training in practice. The injected Gumbel noise promotes stochastic exploration so that different candidates can be selected across iterations, partially compensating for the biased gradient approximation. As the noise scale is annealed during training, the selection gradually becomes deterministic while retaining the benefits of early-stage exploration.

\paragraph{Upsampling}
The coarse predictions $\mathcal{F}^{c}_{pred}$ and $\mathcal{P}^{c}_{pred}$ are fed into a hierarchical upsampling transformer \cite{zhou2022seedformer} to progressively refine the point cloud and recover fine-grained geometric details. 
We employ three upsampling stages, each implemented as $\text{UpTrans}(C_{in,out}=384, C_d=64, K=16, U)$, where the upsampling rates are $U \in \{1, 4, 4\}$. Here, $C_{in}$ denotes both the input and output feature dimensions, $C_d$ is the latent feature dimension of the upsampling transformer, and $K$ is the number of nearest neighbors used in the subtraction-based attention mechanism. Starting from 512 coarse points, the hierarchical upsampling produces the final prediction  $\mathcal{P}^{f}_{pred} \in \mathbb{R}^{8192 \times 3}$.

\section{Metrics Details}
\label{sec:sup_metrics}
\paragraph{Chamfer Distance}
We use the average Chamfer Distance introduces by \cite{fan2017point} to measures the discrepancy between the predicted point cloud $\mathcal{P}$ and the ground truth $\mathcal{G}$ at the point set level. 
For each prediction, the Chamfer Distance (CD) is calculated as:
\begin{equation}
\label{eq:cd}
CD(\mathcal{P}, \mathcal{G}) = 
\frac{1}{|\mathcal{P}|} \sum_{p \in \mathcal{P}} \min_{g \in \mathcal{G}} \lVert p - g \rVert
\;+\;
\frac{1}{|\mathcal{G}|} \sum_{g \in \mathcal{G}} \min_{p \in \mathcal{P}} \lVert g - p \rVert .
\end{equation}
We use the L1-norm of the CD (CD$_{\ell1}$) in the loss function and L2-norm (CD$_{\ell2}$) as evaluation metric.

\paragraph{F-Score}
Following the experiment setups of previous works, we also use the F-Score \cite{tatarchenko2019single} as an extra metric for the evaluation, which is define as:
\begin{equation}
\label{eq:fscore}
\mathrm{F\text{-}Score}(d) = 
\frac{2\, P(d)\, R(d)}{P(d) + R(d)},
\end{equation}
where $P(d)$ and $R(d)$ denote the precision and recall with the threshold of distance $d$, respectively. 
\begin{align}
\label{eq:pr}
P(d) = \frac{1}{|\mathcal{P}|}\sum_{p \in \mathcal{P}}\left[ \min_{g \in \mathcal{G}} \lVert p - g \rVert < d \right], \\
R(d) = \frac{1}{|\mathcal{G}|}\sum_{g \in \mathcal{G}}\left[ \min_{p \in \mathcal{P}} \lVert g - p \rVert < d \right].
\end{align}
We set $d = 0.01$ for evaluation across all datasets, and additionally use $d = 0.005$ for the PFS dataset to better capture small discrepancies between the prediction and the ground truth.

\section{Dataset Details}
\label{sec:sup_data}

\paragraph{ShapeNet-55/34}
The ShapeNet-55 dataset contains 55 object categories with 41,952 training samples and 10,518 testing samples. 
ShapeNet-34 is designed to evaluate model generalization on unseen categories, consisting of 46,765 training samples from 34 categories. 
The test set includes 3,400 shapes from 34 seen categories and 2,305 shapes from 21 unseen categories.
Following \cite{yu2021pointr,zhou2022seedformer,chen2023anchorformer}, 
we generate three levels of incompleteness (simple, moderate, and hard) by randomly preserving 75\%, 50\%, and 25\% of the original points, respectively.
We also evaluate the five most frequent categories (Table, Chair, Airplane, Car, and Sofa) and the five least frequent categories (Birdhouse, Bag, Remote, Keyboard, and Rocket), 
in addition to the average over all 55 categories at each incompleteness level.

\paragraph{ShapeNet-Deform}
To enrich geometric diversity and improve robustness to non-rigid variations, we apply a 3D Gabor noise \cite{lagae2009gabor} deformation to all objects of ShapeNet-55 and keep the same partial points generation method from the ground truth. Gabor noise produces smooth, directionally‐coherent perturbations and is widely used to synthesize natural stochastic patterns. For an input point cloud 
$\mathbf{X} \in \mathbb{R}^{N \times 3}$, the deformation is generated by aggregating multiple randomly sampled 3D Gabor kernels.

\textit{Gabor Kernel}:
A single 3D Gabor kernel is defined over a point $\mathbf{x} \in \mathbb{R}^3$ as:
\begin{equation}
    g(\mathbf{x}) =
    \exp\!\left(
        -\tfrac{1}{2}\left( \tfrac{\mathbf{d}^\top \mathbf{x}}{\sigma} \right)^{\!2}
    \right)
    \,
    \times
    \cos\!\left( 2\pi f\, \mathbf{d}^\top \mathbf{x} + \phi \right),
\end{equation}
where  
$\mathbf{d} \in \mathbb{R}^3$ is a unit direction vector, $f$ is the kernel frequency, $\sigma$ is the Gaussian bandwidth, and $\phi$ is the phase shift.
Given a batched point cloud $\mathbf{X} \in \mathbb{R}^{B \times N \times 3}$, we compute a batch-wise projection  
\begin{equation}
    s = \mathbf{d}^\top \mathbf{X} \in \mathbb{R}^{B \times N},
\end{equation}
and evaluate the Gaussian and sinusoidal components as implemented in our batched kernel:
\begin{equation}
    g(\mathbf{X}) = 
    \exp\!\left( -\tfrac{1}{2}(s/\sigma)^2 \right)
    \odot
    \cos\!\left( 2\pi f s + \phi \right).
\end{equation}

\textit{3D Noise Aggregation}:
For each of the three axes, we sample $K$ random kernels and average them:
\begin{equation}
    \mathbf{N}(\mathbf{X})_{:,:,a}
    =
    \frac{1}{K} \sum_{k=1}^{K}
    g\!\left(\mathbf{X} - \mathbf{o}_k; \mathbf{d}_k, f, \sigma, \phi_k \right),
\end{equation}
where $\mathbf{o}_k$ is a random spatial offset.  
This yields a full 3D noise field $\mathbf{N} \in \mathbb{R}^{B \times N \times 3}$.

\textit{Final Deformation}:
The noisy point cloud is obtained via:
\begin{equation}
    \tilde{\mathbf{X}} = \mathbf{X} + \alpha \, \mathbf{N}(\mathbf{X}),
\end{equation}
where $\alpha$ controls the deformation magnitude.

To enhance geometric variability, all Gabor noise parameters are randomly sampled from specified ranges during training. For evaluation, we instead fix the parameters to the midpoint of their ranges and use a fixed random seed to generate deterministic noise. Our setups are given in \cref{tab:sup_gabor}. The dataset is available via \url{https://github.com/ins-uni-bonn/PQDT}.

\begin{table}[h]
  \caption{Gabor noise parameters of ShapeNet-Deform}
  \centering
  \footnotesize
  \setlength{\tabcolsep}{8pt}
  \begin{tabular}{l|ll}
  \toprule
  Param & Training & Evaluation \\
  \midrule
    $\alpha$ & $\mathcal{U}(0.2,\, 0.6)$ & 0.4 \\
    $K$ & $\{8, 9, \dots, 24\}$ & 16 \\
    $f$ & $\mathcal{U}(1.0,\, 3.0)$ & 2.0 \\
    $\sigma$ & $\mathcal{U}(0.4,\, 0.6)$ & 0.5 \\
    $\phi$ & $\mathcal{U}(0, 2\pi)$ & Seed$(0, 2\pi)$ \\
    $\mathbf{d}_k$ & $\mathcal{U}(\text{unit sphere})$ & Seed$(\text{unit sphere})$ \\
    $\mathbf{o}_k$ & $\mathcal{U}(-5, 5)^3$ & Seed$(-5, 5)^3$\\
  \bottomrule
  \end{tabular}
  \label{tab:sup_gabor}
\end{table}

\paragraph{ShapeNetCar-Occ}
To simulate realistic real-world occlusions around vehicles, we construct the ShapeNetCar-Occ dataset by placing random occluding objects (e.g., traffic signs, pedestrians) around each ShapeNet car category and rendering partial observations via raycasting.
Given a car mesh $\mathcal{M}$ and an occluder mesh $\mathcal{O}$, we first sample a virtual LiDAR viewpoint. For each model instance, we compute the ray from the LiDAR origin toward the car center, determine the first intersection with $\mathcal{M}$, and place the occluder near this intersection point. The occluder is randomly rotated around the vertical axis and slightly shifted along the viewing direction to ensure it lies between the sensor and the car while remaining grounded.
A second raycasting pass uniformly samples $n_{rays}$ viewing directions on the sphere (Fibonacci lattice). Only rays whose directions face the car are used. Intersections with the combined scene $\mathcal{S} = \mathcal{M} \cup \mathcal{O}$ produce the observed point set, while a separate raycasting pass on the clean mesh $\mathcal{M}$ identifies which points originate from occluders. Samples with insufficient occlusion points are resampled. Points outside a rescaled car bounding box are removed, and the final point cloud is subsampled to $N$ points using FPS and perturbed with Gaussian noise.

Since online raycasting–based data generation is computationally expensive, we generate the ShapeNetCar-Occ dataset offline using the three difficulty levels defined in \cref{tab:sup_data_occ}.
\begin{table}[t]
\caption{Parameters of ShapeNet-Occ under 3 difficulty levels}
\centering
\footnotesize
\setlength{\tabcolsep}{8.5pt}
    \begin{tabular}{l|ccc}
    \toprule
    Param & Simple & Moderate & Hard \\
    \midrule
    $\beta$ & 1.1 & 1.2 & 1.3 \\
    $\sigma_{noise}$ & 0.003 & 0.004 & 0.005 \\
    $n_{occ}$ & 10 & 20 & 30 \\
    \bottomrule
    \end{tabular}
    \label{tab:sup_data_occ}
\end{table}

Here, $\beta$ denotes the bounding-box rescaling factor: larger values relax spatial filtering and allow more distant points to be preserved. The Gaussian noise level $\sigma_{noise}$ controls the magnitude of sensor-like perturbations. The threshold $n_{occ}$ specifies the minimum number of occlusion points required for a valid sample, thereby regulating occlusion severity. For each difficulty level, 32 partial point clouds are generated per model from random viewpoints. These settings jointly produce progressively more challenging scenarios and enable controlled evaluation of robustness under partial observations.  The dataset is available via \url{https://github.com/ins-uni-bonn/PQDT}.

\begin{figure}[t]
    \centering
    \includegraphics[width=0.99\linewidth]{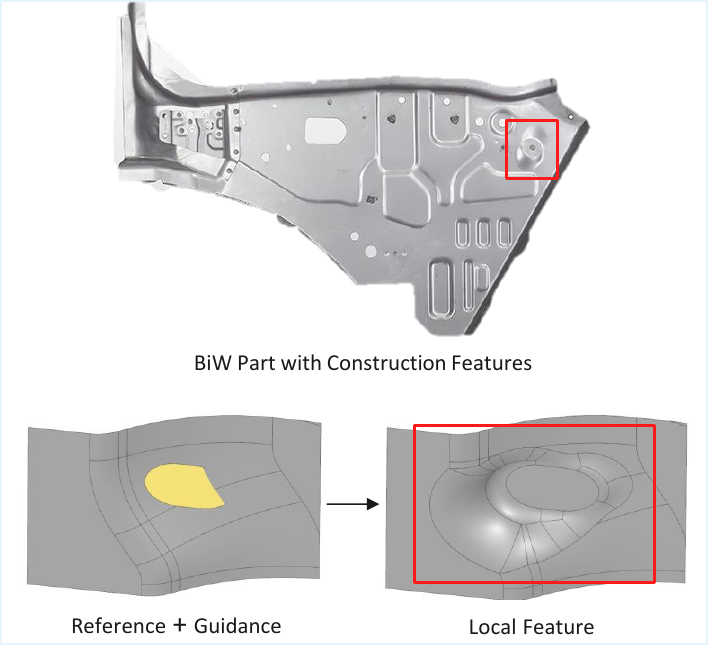}
    \caption{Illustration of PFS usage in an industrial application. The BiW part image (top) is sourced from \cite{skh_biw}.}
    \label{fig:sup_pfs_ill}
\end{figure}

\paragraph{PFS}
We use the patchified freeform surface (PFS) data from Body-in-White (BiW) components in the automotive industry as an additional target domain for our restoration backbone. As illustrated in \cref{fig:sup_pfs_ill}, the dataset is constructed by taking the reference geometry and the guidance geometry (bottom left) as inputs, while the geometry containing local construction features (e.g., embossments for reinforcement, mounting, welding, etc.) serves as the ground truth. The guidance geometry is manually annotated as a simplified cue indicating where local features should be generated. Using CAD software, we capture part snapshots both with and without these local features. The dataset consists of 800 training samples and 80 evaluation samples.

Compared with the previously used datasets, this task exhibits different characteristics: the completion requirement is smaller, while deformation plays a more significant role. Consistency and fine-detail reconstruction become key challenges for this dataset. In practical applications, e.g., reverse engineering, one might start with an incomplete 3D scan and use the guidance geometry to refine the coarse surface. Similarly, this setup enables adding, editing, or inpainting local features on an early-stage CAD model through sketch-based guidance.

Due to the confidential nature of the data and its inclusion of sensitive product-related information, this dataset cannot be made publicly available.

\section{Complexity Analysis}
Benefiting from the dual-stage transformer architecture, our model naturally carries a larger parameter count and higher theoretical computation cost (FLOPs) than most existing approaches. However, this increased capacity directly translates into significantly stronger geometric reasoning and superior completion quality. As shown in \cref{tab:sup_complexity}, PQDT achieves the best reconstruction accuracy on ShapeNet-55 by a clear margin, outperforming all prior baselines across both benchmarks and our newly introduced datasets. Despite its enhanced capability, the model size and computational cost remain well within a practical and efficient range, offering an excellent balance between complexity and the substantial performance gains delivered.
\begin{table}[t]
  \caption{Complexity Analysis}
  \centering
  \footnotesize
  \setlength{\tabcolsep}{7.3pt}
  \begin{tabular}{l|rr|cc}
  \toprule
  Method & Params & FLOPs & CD$_{\ell2}$-55 & CD$_{\ell2}$-34  \\
  \midrule
  FoldingNet \cite{yang2018foldingnet}  & 2.3M & 27.6G & 3.12 & 3.62 \\
  PCN \cite{yuan2018pcn}  & 5.0M & 15.3G & 2.66 & 3.85 \\
  GRNet \cite{xie2020grnet}  & 73.2M & 40.4G & 1.97 & 2.99 \\
  PoinTr \cite{yu2021pointr}  & 33.0M & 8.9G & 1.09 & 2.05 \\
  SeedFormer \cite{zhou2022seedformer}  & 3.2M & 10.3G & 0.92 & 1.34 \\
  AdaPoinTr \cite{Yu2023AdaPoinTrDP}  & 32.5M & 18.1G & 0.81 & 1.23 \\
  ProxyFormer \cite{li2023proxyformer} & 12.2M & 9.9G & 0.93 & 1.42 \\
  AnchorFormer \cite{chen2023anchorformer}  & 30.5M & 8.1G & 0.76 & 1.19 \\
  \midrule
  PQDT  & 64.2M & 60.3G & \textbf{0.68} & \textbf{0.99} \\
  \bottomrule
  \end{tabular}
  \label{tab:sup_complexity}
\end{table}

\section{Additional Ablation Study}
\label{sec:sup_ab}
To evaluate the effectiveness of our model design, we conduct a comprehensive ablation study, as summarized in \cref{tab:ablation}, focusing on the submodules introduced in \cref{sec:method}. 
We use a vanilla Transformer for point clouds as the baseline (Model A) and analyze performance differences across three key aspects: query generation, seed prediction, and positional encoding. 
In each experiment, we modify only the component of interest while keeping all other settings consistent with the best-performing configuration.

\paragraph{Query Generation}
Starting from the single-stage variant (Model B), we first incorporate the pseudo-query mechanism (Model C) to examine the impact of introducing the second stage in the transformer. Importantly, this transition does not involve adding extra transformer layers; instead, it restructures the query formulation within the existing architecture. As shown in \cref{tab:ablation}, the parameter count increases only marginally from 55.0M to 57.8M, indicating that the performance gains are not simply due to increased model capacity.
Despite this minimal overhead, the addition of pseudo-queries leads to consistent improvements in both CD and F1-Score, clearly demonstrating the effectiveness of our dual-stage design. This suggests that the performance boost stems from better feature refinement and query representation rather than scaling the model size.
Further introducing the dynamic query selection (DQS) module (PQDT) yields additional gains, 
achieving a CD of 0.82 and an F1-Score of 0.261.

\paragraph{Seed Prediction}
We compare the commonly used linear projection from global features (Model D) with our query-based DETR-style decoding scheme, 
which achieves a significant improvement, particularly in F1-Score. 
Using partial input initialization (Model E) performs slightly worse than the spherical initialization adopted in our final model, 
suggesting that the latter provides a more stable prior for query generation.

\paragraph{Positional Encoding}
Model F encodes spatial information using only the point coordinates, 
while Model G incorporates local context through $k$NN-based neighborhood aggregation. 
The latter slightly decreases CD performance (by 0.02), likely due to overfitting. 
In contrast, our geometric-embedding achieves the best overall results, 
highlighting the advantage of our geometry-aware positional design.

\begin{table}[t]
\caption{Ablation study on ShapeNetCar-Occ. 
We report the evaluation results and model sizes with different model designs including 
single query generation (Query), auxiliary pseudo-query generation (Pseudo-Query), Dynamic Query Selection (DQS), 
linear seed projection (Linear Proj.), query-based decoder with partial points 
and spherical points as initialization (P/S-Q Dec.), 
coordinates encoding (CE), $k$NN feature grouping ($k$NN), and geometric-embedding (GE).
}
\centering
\scriptsize
\setlength{\tabcolsep}{3pt} 
    \begin{tabular}{cc|ccc|cc}
    \toprule
    Category & Model & \multicolumn{3}{c|}{Setups} & CD$_{\ell2}$ ($\downarrow$) & F1 ($\uparrow$) \\
    \midrule
    Baseline & A & \multicolumn{3}{c|}{Query + Linear Proj. + SA} & 0.857 & 0.243 \\
    \midrule
    \multirow{4}{*}{\shortstack{Query\\Generation}} &  & Query & Pseudo-Query & DQS & CD$_{\ell2}$ ($\downarrow$) & F1 ($\uparrow$) \\
    \cmidrule(l){2-7}
     & B & \checkmark &  &  & 0.838 & 0.256 \\
     & C & \checkmark & \checkmark &  & 0.827 & 0.260 \\
     & PQDT & \checkmark & \checkmark & \checkmark & 0.818 & 0.261 \\
    \midrule
    \multirow{4}{*}{\shortstack{Seed\\Prediction}} &  & Linear Proj. & P-Q Dec. & S-Q Dec. & CD$_{\ell2}$ ($\downarrow$) & F1 ($\uparrow$) \\
    \cmidrule(l){2-7}
     & D & \checkmark &  &  & 0.851 & 0.235 \\
     & E &  & \checkmark &  & 0.841 & 0.260 \\
     & PQDT &  &  & \checkmark & 0.818 & 0.261 \\
    \midrule
    \multirow{4}{*}{\shortstack{Positional\\Encoding}} & & CE & $k$NN & GE & CD$_{\ell2}$ ($\downarrow$) & F1 ($\uparrow$) \\
    \cmidrule(l){2-7}
     & F & \checkmark &  &  & 0.850 & 0.248 \\
     & G & \checkmark & \checkmark &  & 0.865 & 0.251 \\
     & PQDT & \checkmark &  & \checkmark & 0.818 & 0.261 \\
    \bottomrule
    \end{tabular}
    \setlength{\tabcolsep}{3.78pt} 
    \begin{tabular}{c|cccccccc}
      \toprule
      Model & A & B & C & D & E & F & G & PQDT \\
      \midrule
      Params & 44.9M & 55.0M & 57.8M & 58.6M & 64.2M & 63.1M & 63.1M & 64.2M \\
      \bottomrule
      \end{tabular}
    \label{tab:ablation}
\end{table}

We also investigate the component level ablation study for the DQS and query-based seed generation on PFS in \cref{tab:sup_ab_gumbel,tab:sup_ab_detr}.

\paragraph{Gumbel Noise Perturbation}
We vary the scaling factor of the Gumbel noise added to the query scores in the DQS module to study its effect on performance. A larger noise scale introduces more randomness in query selection, which can strengthen the regularization effect; however, if the noise becomes too large, performance degrades. Conversely, using a very small scale (i.e., effectively no noise) also harms performance. In our setup, we use a noise scale of 1.0.
\begin{table}[h]
  \caption{Ablation study on Gumbel-Top-$k$ of dynamic query selection module on ShapeNetCar-Occ.}
  \centering
  \footnotesize
  \setlength{\tabcolsep}{6.9pt}
  \begin{tabular}{l|cc}
  \toprule
  Method & CD$_{\ell2}$ ($\downarrow$) & F1 ($\uparrow$) \\
  \midrule
  w/o noise         & 0.833 & 0.253 \\
  noise scale = 0.5 & 0.825 & 0.256 \\
  noise scale = 1.0 & 0.818 & 0.261 \\
  noise scale = 2.0 & 0.834 & 0.247 \\
  \bottomrule
  \end{tabular}
  \label{tab:sup_ab_gumbel}
\end{table}

\begin{figure}[t]
    \centering
    \includegraphics[width=0.9\linewidth]{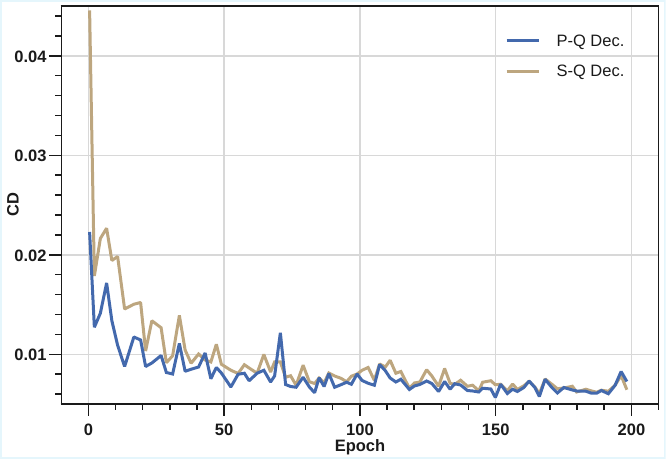}
    \caption{Training loss under different initializations of query-based seed generation on PFS.
    }
    \label{fig:sup_psq}
\end{figure}

\paragraph{Initialization of Seed Generation}
Beyond the comparison with Model E in \cref{tab:ablation} on ShapeNetCar-Occ, we also investigate how different initializations of the query-based decoder influence seed generation in the first stage of the transformer. We evaluate partial-initialized and spherical-initialized queries (P/S-Q Dec.) on the PFS dataset. Because the input and ground-truth shapes in PFS share a substantial amount of common geometry, using the partial input as queries should theoretically yield better performance. As shown in \cref{tab:sup_ab_detr}, both variants achieve similar CD scores, but P-Q Dec. attains a higher F-Score, and \cref{fig:sup_psq} further shows that it converges faster.

\begin{table}[t]
  \caption{Ablation study on query-based seed generation with partial points 
    and spherical points as initialization (P/S-Q Dec.) on PFS.}
  \centering
  \footnotesize
  \setlength{\tabcolsep}{6.9pt}
  \begin{tabular}{l|ccc}
  \toprule
  Method & CD$_{\ell2}$ ($\downarrow$) & F1 ($\uparrow$) & F0.5 ($\uparrow$) \\
  \midrule
  P-Q Dec.  & 0.17 & 0.847 & 0.327 \\
  S-Q Dec.  & 0.16 & 0.834 & 0.311 \\
  \bottomrule
  \end{tabular}
  \label{tab:sup_ab_detr}
\end{table}

In all experiments across the datasets mentioned in the paper, we adopt spherical initialization to demonstrate the generality of our model. However, it is also possible to tune the query initialization, such as using partial input, to better suit specific tasks or datasets.

\section{Additional Evaluation Results}
\label{sec:sup_eval}
\begin{table*}[t]
  \caption{Results of our method and state-of-the-art methods on ShapeNet-55. 
  We report the detailed results for each method on 10 categories and the overall results on 55 categories 
  for three difficulty degrees. We use CD-S, CD-M and CD-H to represent the CD results under the 
  Simple, Moderate and Hard settings. We also provide results under the F-Score@1\% metric.}
  \centering
  \footnotesize
  \setlength{\tabcolsep}{5.5pt}
  \begin{tabular}{l ccccc|ccccc|ccc|cc}
  \toprule
  Method & Tab. & Cha. & Pla. & Car & Sof. & Bir. & Bag & Rem. & Key. & Roc. 
  & CD-S & CD-M & CD-H & CD$_{\ell2}$ ($\downarrow$) & F1 ($\uparrow$) \\
  \midrule
  FoldingNet \cite{yang2018foldingnet} & 2.53 & 2.81 & 1.43 & 1.98 & 2.48 & 4.71 & 2.79 & 1.44 & 1.24 & 1.48 & 2.67 & 2.66 & 4.05 & 3.12 & 0.082 \\
  PCN \cite{yuan2018pcn} & 2.13 & 2.29 & 1.02 & 1.85 & 2.06 & 4.50 & 2.86 & 1.32 & 0.89 & 1.32 & 1.94 & 1.96 & 4.08 & 2.66 & 0.133 \\
  GRNet \cite{xie2020grnet} & 1.63 & 1.88 & 1.02 & 1.62 & 1.72 & 2.97 & 2.06 & 1.09 & 0.89 & 1.03 & 1.35 & 1.71 & 2.85 & 1.97 & 0.238 \\
  PoinTr \cite{yu2021pointr} & 0.81 & 0.95 & 0.44 & 0.91 & 0.79 & 1.86 & 0.93 & 0.53 & 0.38 & 0.57 & 0.58 & 0.88 & 1.79 & 1.09 & 0.464 \\
  SnowflakeNet \cite{xiang2021snowflakenet} & 0.98 & 1.12 & 0.54 & 0.98 & 1.02 & 1.93 & 1.08 & 0.57 & 0.48 & 0.61 & 0.70 & 1.06 & 1.96 & 1.24 & 0.398 \\
  SeedFormer \cite{zhou2022seedformer} & 0.72 & 0.81 & 0.40 & 0.89 & 0.71 & 1.51 & 0.79 & 0.46 & 0.36 & 0.50 & 0.50 & 0.77 & 1.49 & 0.92 & 0.472 \\
  AdaPoinTr \cite{Yu2023AdaPoinTrDP} & 0.62 & 0.69 & 0.33 & 0.81 & 0.63 & 1.33 & 0.68 & 0.38 & 0.33 & \textbf{0.34} & 0.49 & 0.69 & 1.24 & 0.81 & 0.503 \\
  ProxyFormer \cite{li2023proxyformer} & 0.70 & 0.83 & 0.34 & 0.78 & 0.69 & 1.57 & 0.79 & 0.36 & 0.34 & 0.46 & 0.49 & 0.75 & 1.55 & 0.93 & 0.483 \\
  T-CorresNet \cite{duan2024t} & 0.60 & 0.68 & \textbf{0.32} & - & - & 1.17 & \textbf{0.60} & - & 0.27 & - & 0.50 & 0.68 & 1.23 & 0.80 & 0.485 \\
  AnchorFormer \cite{chen2023anchorformer} & 0.58 & 0.67 & 0.33 & 0.69 & 0.58 & 1.35 & 0.64 & 0.36 & 0.27 & 0.42 & 0.41 & 0.61 & 1.26 & 0.76 & 0.558 \\
  \midrule
  \textbf{PQDT} & \textbf{0.52} & \textbf{0.60} & \textbf{0.32} & \textbf{0.65} & \textbf{0.51} & \textbf{1.13} 
    & \textbf{0.60} & \textbf{0.31} & \textbf{0.24} & 0.41 & \textbf{0.34} & \textbf{0.55} & \textbf{1.13} & \textbf{0.68} & \textbf{0.570} \\
  \bottomrule
  \end{tabular}
  \label{tab:sup_shapenet55}
\end{table*}

In \cref{tab:sup_shapenet55}, we report results for the five most frequent categories (Table, Chair, Airplane, Car, and Sofa) and the five least frequent categories (Birdhouse, Bag, Remote, Keyboard, and Rocket), along with the average CD$_{\ell2}$ and F1-Score. These metrics are compared against the baselines under three levels of incompleteness. In \cref{tab:sup_shapenet_55_deform_cat}, we further provide categorical results for all 55 categories on ShapeNet-55 and ShapeNet-Deform. In \cref{tab:sup_shapenet_34_cat}, we present results on the 22 unseen categories from ShapeNet-34. Finally, in \cref{tab:sup_shapenet_occ}, we report detailed comparisons with baselines across the three difficulty levels of ShapeNetCar-Occ.

\begin{table*}[h]
    \caption{Categorical results of our method on ShapeNet-55 and ShapeNet-Deform. 
      We report the detailed results under CD$_{\ell2}$ and F-Score@1\% metric for Simple, Moderate and Hard settings.}
    \centering
    \footnotesize
    \setlength{\tabcolsep}{6pt}
    \begin{tabular}{lcccccc | cccccc}
        \toprule
        \multirow{4}{*}{Category} 
        & \multicolumn{6}{c}{\textbf{ShapeNet-55}} 
        & \multicolumn{6}{c}{\textbf{ShapeNet-Deform}} \\ 
        \cmidrule(lr){2-7}\cmidrule(lr){8-13}
        & \multicolumn{2}{c}{Simple} & \multicolumn{2}{c}{Moderate} & \multicolumn{2}{c}{Hard}
        & \multicolumn{2}{c}{Simple} & \multicolumn{2}{c}{Moderate} & \multicolumn{2}{c}{Hard} \\ 
        \cmidrule(lr){2-3}\cmidrule(lr){4-5}\cmidrule(lr){6-7}
        \cmidrule(lr){8-9}\cmidrule(lr){10-11}\cmidrule(lr){12-13}
        & CD$_{\ell2}$ ($\downarrow$) & F1 ($\uparrow$)
        & CD$_{\ell2}$ ($\downarrow$) & F1 ($\uparrow$)
        & CD$_{\ell2}$ ($\downarrow$) & F1 ($\uparrow$)
        & CD$_{\ell2}$ ($\downarrow$) & F1 ($\uparrow$)
        & CD$_{\ell2}$ ($\downarrow$) & F1 ($\uparrow$)
        & CD$_{\ell2}$ ($\downarrow$) & F1 ($\uparrow$) \\
        \midrule
        airplane & 0.18 & 0.765 & 0.26 & 0.743 & 0.51 & 0.627 & 0.36 & 0.519 & 0.45 & 0.498 & 0.70 & 0.433 \\ 
        trash bin & 0.49 & 0.472 & 0.77 & 0.461 & 1.48 & 0.413 & 0.83 & 0.159 & 1.09 & 0.151 & 2.13 & 0.126 \\ 
        bag & 0.32 & 0.590 & 0.52 & 0.576 & 0.95 & 0.492 & 0.63 & 0.270 & 0.82 & 0.252 & 1.30 & 0.210 \\ 
        basket & 0.46 & 0.503 & 0.56 & 0.491 & 1.10 & 0.434 & 0.76 & 0.193 & 0.87 & 0.180 & 1.68 & 0.146 \\ 
        bathtub & 0.37 & 0.549 & 0.58 & 0.533 & 1.07 & 0.467 & 0.69 & 0.237 & 0.89 & 0.219 & 1.48 & 0.177 \\ 
        bed & 0.42 & 0.544 & 0.65 & 0.532 & 1.35 & 0.466 & 0.78 & 0.226 & 1.01 & 0.207 & 1.79 & 0.174 \\ 
        bench & 0.22 & 0.651 & 0.32 & 0.653 & 0.65 & 0.587 & 0.42 & 0.399 & 0.50 & 0.372 & 0.86 & 0.324 \\ 
        birdhouse & 0.55 & 0.517 & 0.94 & 0.500 & 1.90 & 0.433 & 0.98 & 0.177 & 1.37 & 0.164 & 2.60 & 0.141 \\ 
        bookshelf & 0.40 & 0.519 & 0.62 & 0.509 & 1.22 & 0.447 & 0.71 & 0.222 & 0.92 & 0.210 & 1.59 & 0.181 \\ 
        bottle & 0.22 & 0.650 & 0.47 & 0.615 & 0.98 & 0.507 & 0.47 & 0.365 & 0.75 & 0.331 & 1.41 & 0.246 \\ 
        bowl & 0.39 & 0.502 & 0.46 & 0.493 & 0.85 & 0.433 & 0.63 & 0.195 & 0.73 & 0.179 & 1.18 & 0.151 \\ 
        bus & 0.28 & 0.588 & 0.40 & 0.584 & 0.58 & 0.533 & 0.54 & 0.321 & 0.64 & 0.299 & 0.85 & 0.245 \\ 
        cabinet & 0.38 & 0.499 & 0.47 & 0.495 & 0.80 & 0.454 & 0.64 & 0.209 & 0.73 & 0.199 & 1.05 & 0.175 \\ 
        camera & 0.56 & 0.523 & 1.26 & 0.504 & 2.63 & 0.430 & 1.09 & 0.171 & 1.76 & 0.164 & 3.39 & 0.140 \\ 
        can & 0.38 & 0.515 & 0.69 & 0.501 & 1.48 & 0.439 & 0.65 & 0.248 & 1.00 & 0.235 & 2.30 & 0.195 \\ 
        cap & 0.33 & 0.572 & 0.32 & 0.554 & 0.67 & 0.460 & 0.47 & 0.281 & 0.59 & 0.252 & 1.00 & 0.210 \\ 
        car & 0.42 & 0.489 & 0.62 & 0.475 & 0.93 & 0.424 & 0.89 & 0.142 & 1.01 & 0.137 & 1.28 & 0.123 \\ 
        cellphone & 0.21 & 0.626 & 0.26 & 0.620 & 0.37 & 0.556 & 0.36 & 0.431 & 0.41 & 0.416 & 0.56 & 0.339 \\ 
        chair & 0.28 & 0.610 & 0.47 & 0.591 & 1.06 & 0.492 & 0.57 & 0.291 & 0.78 & 0.263 & 1.41 & 0.226 \\ 
        clock & 0.37 & 0.559 & 0.56 & 0.547 & 1.05 & 0.484 & 0.64 & 0.280 & 0.79 & 0.267 & 1.34 & 0.227 \\ 
        keyboard & 0.18 & 0.650 & 0.24 & 0.657 & 0.31 & 0.596 & 0.36 & 0.445 & 0.42 & 0.408 & 0.60 & 0.374 \\ 
        dishwasher & 0.38 & 0.503 & 0.47 & 0.500 & 1.03 & 0.450 & 0.63 & 0.223 & 0.75 & 0.205 & 1.39 & 0.180 \\ 
        display & 0.27 & 0.583 & 0.43 & 0.579 & 0.84 & 0.518 & 0.51 & 0.317 & 0.65 & 0.295 & 1.11 & 0.260 \\ 
        earphone & 0.45 & 0.600 & 0.77 & 0.560 & 2.00 & 0.445 & 0.95 & 0.225 & 1.17 & 0.216 & 3.06 & 0.186 \\ 
        faucet & 0.38 & 0.734 & 0.82 & 0.658 & 1.94 & 0.483 & 0.71 & 0.391 & 1.14 & 0.353 & 2.52 & 0.276 \\ 
        file cabinet & 0.42 & 0.502 & 0.54 & 0.498 & 1.14 & 0.448 & 0.71 & 0.209 & 0.86 & 0.201 & 1.54 & 0.176 \\ 
        guitar & 0.09 & 0.929 & 0.14 & 0.880 & 0.25 & 0.794 & 0.19 & 0.793 & 0.25 & 0.775 & 0.37 & 0.715 \\ 
        helmet & 0.59 & 0.501 & 1.17 & 0.480 & 2.92 & 0.410 & 1.13 & 0.141 & 1.70 & 0.128 & 3.38 & 0.107 \\ 
        jar & 0.44 & 0.537 & 0.79 & 0.512 & 1.79 & 0.433 & 0.81 & 0.211 & 1.20 & 0.193 & 2.51 & 0.156 \\ 
        knife & 0.10 & 0.926 & 0.21 & 0.832 & 0.36 & 0.700 & 0.24 & 0.766 & 0.33 & 0.724 & 0.51 & 0.644 \\ 
        lamp & 0.31 & 0.767 & 0.81 & 0.700 & 2.06 & 0.554 & 0.64 & 0.443 & 1.17 & 0.406 & 2.88 & 0.339 \\ 
        laptop & 0.21 & 0.601 & 0.24 & 0.599 & 0.39 & 0.534 & 0.37 & 0.371 & 0.41 & 0.349 & 0.57 & 0.304 \\ 
        loudspeaker & 0.46 & 0.519 & 0.74 & 0.509 & 1.42 & 0.451 & 0.82 & 0.217 & 1.09 & 0.201 & 1.82 & 0.171 \\ 
        mailbox & 0.19 & 0.758 & 0.47 & 0.703 & 1.72 & 0.548 & 0.45 & 0.463 & 0.82 & 0.404 & 2.28 & 0.334 \\ 
        microphone & 0.39 & 0.790 & 0.93 & 0.707 & 2.32 & 0.546 & 0.65 & 0.482 & 1.24 & 0.423 & 3.58 & 0.331 \\ 
        microwaves & 0.42 & 0.503 & 0.54 & 0.500 & 1.21 & 0.455 & 0.71 & 0.208 & 0.90 & 0.183 & 1.54 & 0.165 \\ 
        motorbike & 0.45 & 0.535 & 0.72 & 0.513 & 1.16 & 0.418 & 0.97 & 0.178 & 1.14 & 0.176 & 1.46 & 0.158 \\ 
        mug & 0.54 & 0.473 & 0.74 & 0.462 & 1.53 & 0.413 & 0.85 & 0.165 & 1.08 & 0.155 & 2.08 & 0.127 \\ 
        piano & 0.44 & 0.543 & 0.59 & 0.540 & 1.19 & 0.484 & 0.73 & 0.249 & 0.91 & 0.230 & 1.45 & 0.190 \\ 
        pillow & 0.35 & 0.573 & 0.50 & 0.543 & 0.97 & 0.448 & 0.72 & 0.206 & 0.87 & 0.183 & 1.38 & 0.143 \\ 
        pistol & 0.26 & 0.701 & 0.42 & 0.657 & 0.70 & 0.528 & 0.53 & 0.394 & 0.66 & 0.380 & 0.93 & 0.339 \\ 
        flowerpot & 0.59 & 0.522 & 0.91 & 0.501 & 1.74 & 0.427 & 1.06 & 0.179 & 1.33 & 0.167 & 2.28 & 0.141 \\ 
        printer & 0.42 & 0.533 & 0.75 & 0.522 & 1.61 & 0.455 & 0.84 & 0.200 & 1.19 & 0.187 & 2.21 & 0.161 \\ 
        remote & 0.19 & 0.666 & 0.31 & 0.651 & 0.44 & 0.564 & 0.39 & 0.452 & 0.50 & 0.437 & 0.69 & 0.336 \\ 
        rifle & 0.19 & 0.855 & 0.31 & 0.787 & 0.54 & 0.669 & 0.37 & 0.639 & 0.51 & 0.610 & 0.75 & 0.548 \\ 
        rocket & 0.14 & 0.862 & 0.37 & 0.789 & 0.72 & 0.687 & 0.35 & 0.611 & 0.61 & 0.556 & 1.00 & 0.474 \\ 
        skateboard & 0.16 & 0.737 & 0.28 & 0.714 & 0.46 & 0.625 & 0.33 & 0.504 & 0.41 & 0.479 & 0.56 & 0.409 \\ 
        sofa & 0.34 & 0.544 & 0.45 & 0.539 & 0.75 & 0.489 & 0.66 & 0.236 & 0.74 & 0.218 & 1.09 & 0.182 \\ 
        stove & 0.41 & 0.544 & 0.63 & 0.531 & 1.18 & 0.465 & 0.73 & 0.234 & 0.97 & 0.217 & 1.55 & 0.191 \\ 
        table & 0.29 & 0.604 & 0.41 & 0.601 & 0.86 & 0.542 & 0.49 & 0.356 & 0.61 & 0.336 & 1.11 & 0.286 \\ 
        telephone & 0.21 & 0.625 & 0.26 & 0.623 & 0.39 & 0.562 & 0.36 & 0.436 & 0.42 & 0.425 & 0.57 & 0.352 \\ 
        tower & 0.31 & 0.663 & 0.58 & 0.617 & 1.22 & 0.491 & 0.62 & 0.364 & 0.92 & 0.331 & 1.87 & 0.258 \\ 
        train & 0.31 & 0.617 & 0.46 & 0.607 & 0.76 & 0.533 & 0.61 & 0.317 & 0.76 & 0.303 & 1.10 & 0.253 \\ 
        watercraft & 0.24 & 0.717 & 0.42 & 0.678 & 0.72 & 0.566 & 0.55 & 0.376 & 0.72 & 0.345 & 1.08 & 0.283 \\ 
        washer & 0.44 & 0.494 & 0.62 & 0.487 & 1.70 & 0.440 & 0.79 & 0.197 & 1.00 & 0.179 & 1.88 & 0.158 \\ 
        \midrule
        mean & 0.34 & 0.608 & 0.55 & 0.586 & 1.13 & 0.505 & 0.63 & 0.319 & 0.85 & 0.298 & 1.54 & 0.253 \\ 
        \bottomrule
    \end{tabular}
    \label{tab:sup_shapenet_55_deform_cat}
\end{table*}

\begin{table}[h]
    \caption{Categorical results of our method on 22 unseen objects of ShapeNet-34. 
      We report the detailed results under CD$_{\ell2}$ and F-Score@1\% metric for Simple, Moderate and Hard settings.}
    \centering
    \footnotesize
    \setlength{\tabcolsep}{3.5pt}
    \begin{tabular}{lcccccc}
        \toprule
        \multirow{4}{*}{Category} & \multicolumn{6}{c}{\textbf{ShapeNet-34}} \\
        \cmidrule(lr){2-7}
        & \multicolumn{2}{c}{Simple} & \multicolumn{2}{c}{Moderate} & \multicolumn{2}{c}{Hard} \\
        \cmidrule(lr){2-3}\cmidrule(lr){4-5}\cmidrule(lr){6-7}
        & CD$_{\ell2}$ ($\downarrow$) & F1 ($\uparrow$) 
        & CD$_{\ell2}$ ($\downarrow$) & F1 ($\uparrow$) 
        & CD$_{\ell2}$ ($\downarrow$) & F1 ($\uparrow$) \\ 
        \midrule
        bag & 0.37 & 0.579 & 0.61 & 0.568 & 1.12 & 0.485 \\ 
        basket & 0.41 & 0.516 & 0.60 & 0.505 & 1.50 & 0.438 \\ 
        birdhouse & 0.54 & 0.545 & 0.91 & 0.527 & 1.90 & 0.452 \\ 
        bowl & 0.42 & 0.499 & 0.53 & 0.488 & 1.11 & 0.426 \\ 
        camera & 0.62 & 0.528 & 1.30 & 0.507 & 2.98 & 0.432 \\ 
        can & 0.38 & 0.508 & 0.61 & 0.496 & 1.23 & 0.436 \\ 
        cap & 0.33 & 0.569 & 0.94 & 0.532 & 3.66 & 0.426 \\ 
        keyboard & 0.22 & 0.631 & 0.27 & 0.634 & 0.38 & 0.576 \\ 
        dishwasher & 0.42 & 0.495 & 0.55 & 0.491 & 1.13 & 0.443 \\ 
        earphone & 0.56 & 0.604 & 1.48 & 0.555 & 4.83 & 0.437 \\ 
        helmet & 0.69 & 0.494 & 1.76 & 0.471 & 4.21 & 0.402 \\ 
        mailbox & 0.32 & 0.721 & 0.67 & 0.670 & 1.64 & 0.531 \\ 
        microphone & 0.53 & 0.803 & 1.16 & 0.723 & 3.04 & 0.551 \\ 
        microwaves & 0.44 & 0.495 & 0.58 & 0.492 & 1.23 & 0.447 \\ 
        pillow & 0.32 & 0.581 & 0.49 & 0.552 & 1.08 & 0.455 \\ 
        printer & 0.47 & 0.522 & 0.83 & 0.512 & 1.91 & 0.449 \\ 
        remote & 0.19 & 0.670 & 0.30 & 0.660 & 0.42 & 0.583 \\ 
        rocket & 0.20 & 0.854 & 0.41 & 0.779 & 0.83 & 0.652 \\ 
        skateboard & 0.21 & 0.724 & 0.49 & 0.683 & 0.84 & 0.535 \\ 
        tower & 0.34 & 0.655 & 0.66 & 0.612 & 1.47 & 0.488 \\ 
        washer & 0.43 & 0.497 & 0.61 & 0.491 & 1.54 & 0.443 \\ 
        \midrule
        mean & 0.40 & 0.595 & 0.75 & 0.569 & 1.81 & 0.480 \\ 
        \bottomrule
    \end{tabular}
    \label{tab:sup_shapenet_34_cat}
\end{table}

\begin{table}[h]
  \caption{Results of our method and state-of-the-art methods on ShapeNetCar-Occ. 
  We report the detailed results for each method on Car category for three difficulty degrees. 
  We use CD-S, CD-M and CD-H to represent the CD results under the 
  Simple, Moderate and Hard settings. We also provide results under the F-Score@1\% metric.
  }
  \centering
  \footnotesize
  \setlength{\tabcolsep}{5.5pt}
  \begin{tabular}{l ccc|cc}
  \toprule
  Method & CD-S & CD-M & CD-H & CD$_{\ell2}$ ($\downarrow$) & F1 ($\uparrow$) \\
  \midrule
  PCN \cite{yuan2018pcn} & 1.41 & 1.41 & 1.43 & 1.42 & 0.125 \\
  PoinTr \cite{yu2021pointr} & 1.06 & 1.07 & 1.07 & 1.07 & 0.191 \\
  SnowflakeNet \cite{xiang2021snowflakenet} & 0.93 & 0.94 & 0.96 & 0.94 & 0.22 \\
  SeedFormer \cite{zhou2022seedformer} & 0.95 & 0.97 & 0.99 & 0.97 & 0.231 \\
  AdaPoinTr \cite{Yu2023AdaPoinTrDP} & 0.84 & 0.84 & 0.86 & 0.85 & \textbf{0.251} \\
  AnchorFormer \cite{chen2023anchorformer} & 0.92 & 0.93 & 0.95 & 0.93 & 0.225 \\
  \midrule
  \textbf{PQDT} & \textbf{0.81} & \textbf{0.82} & \textbf{0.82} & \textbf{0.82} & \textbf{0.261} \\
  \bottomrule
  \end{tabular}
  \label{tab:sup_shapenet_occ}
\end{table}

\section{Qualitative Samples}
\label{sec:sup_sample}
\begin{figure*}[t]
    \centering
    \includegraphics[width=0.99\linewidth]{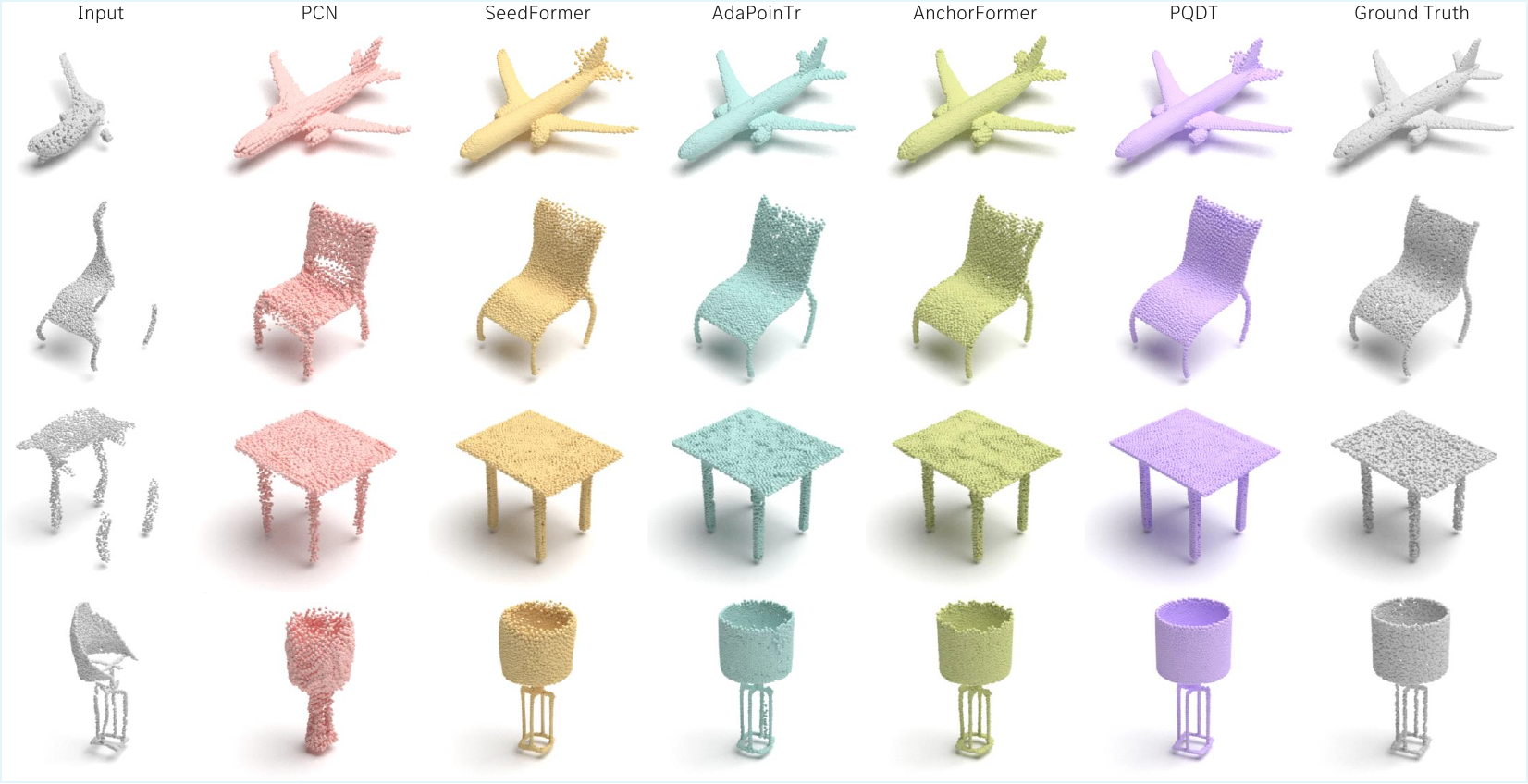}
    \caption{Qualitative evaluation on ShapeNet-Deform
    }
    \label{fig:sup_vis_deform}
\end{figure*}

\begin{figure*}[h]
    \centering
    \includegraphics[width=0.99\linewidth]{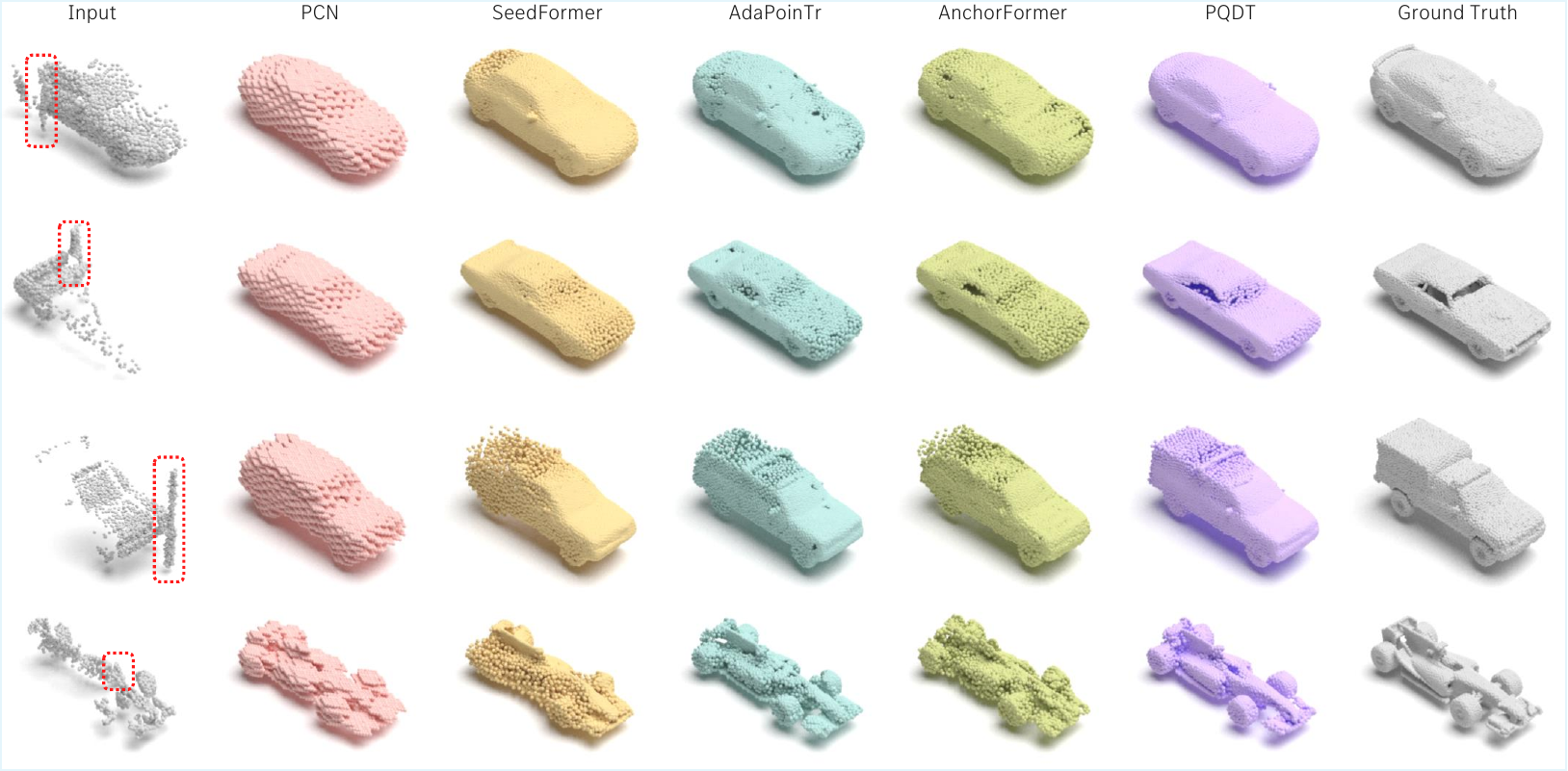}
    \caption{Qualitative evaluation on ShapeNetCar-Occ. Red boxes indicate occluding objects.
    }
    \label{fig:sup_vis_occ}
\end{figure*}

\begin{figure}[h]
    \centering
    \includegraphics[width=0.99\linewidth]{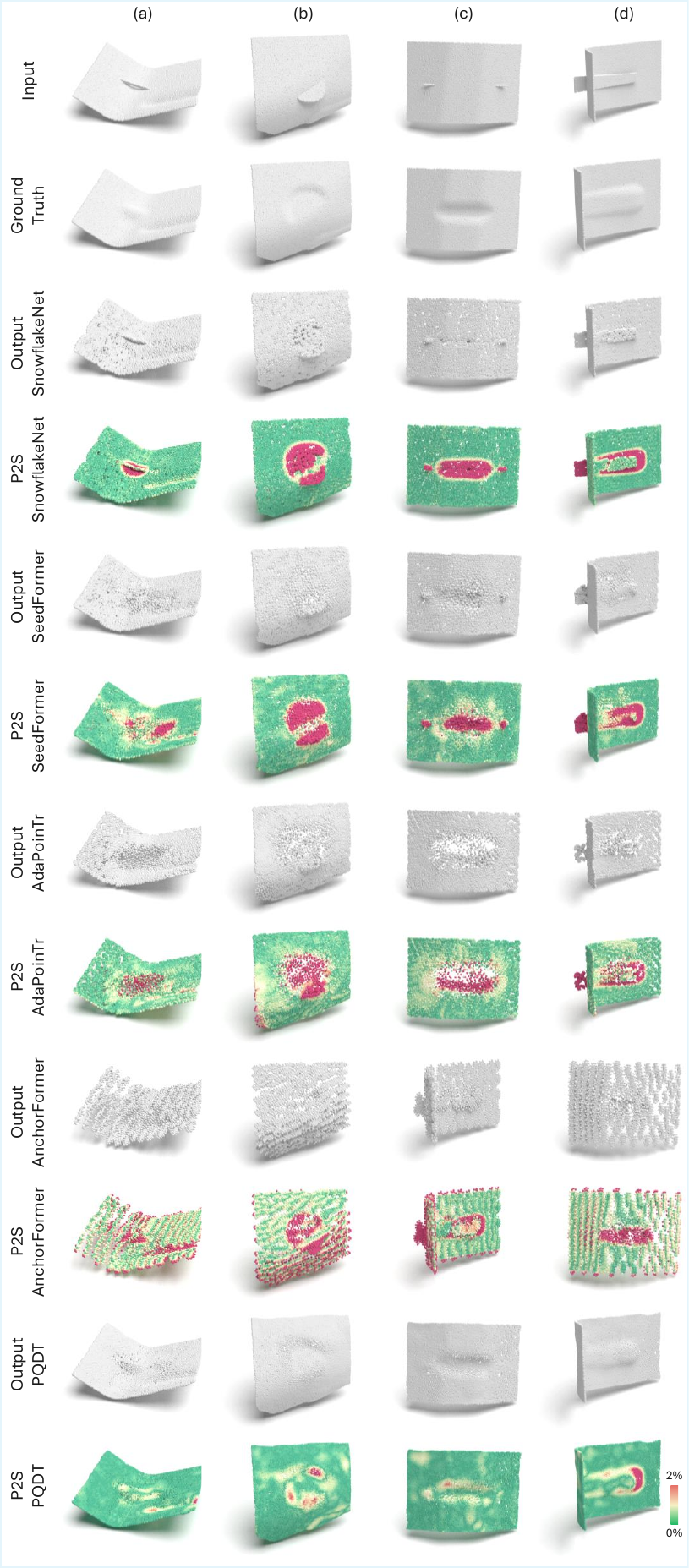}
    \caption{
    Qualitative evaluation on PFS. Colors indicate the point-to-surface (P2S) distance between the generated point clouds and the ground-truth mesh, with the maximum value (red) clamped to 2\% of the radius of the object’s bounding sphere. Examples (a)–(d) show BiW patches containing local construction features such as reinforcement or mounting regions, comparing our method against baseline approaches.
    }
    \label{fig:sup_vis_pfs}
\end{figure}

We provide additional qualitative results for PQDT on ShapeNet-Deform in \cref{fig:sup_vis_deform}, ShapeNetCar-Occ in \cref{fig:sup_vis_occ}, and PFS in \cref{fig:sup_vis_pfs}.


{
    %\clearpage
    \small
    \bibliographystyle{ieeenat_fullname}
    \bibliography{main}
}

\end{document}